\documentclass{article} 
\usepackage{iclr2025_conference,times}

\usepackage{wrapfig}
\usepackage{lipsum} 
\usepackage{wrapfig}

\definecolor{hjn}{RGB}{0,0,0}
\definecolor{why}{RGB}{0, 0, 0}
\definecolor{check}{RGB}{255, 0, 0}

\usepackage{amsmath,amsfonts,bm}
\usepackage{multicol}

\def\eqref#1{equation~\ref{#1}}

\def\1{\bm{1}}

\DeclareMathAlphabet{\mathsfit}{\encodingdefault}{\sfdefault}{m}{sl}
\SetMathAlphabet{\mathsfit}{bold}{\encodingdefault}{\sfdefault}{bx}{n}

\usepackage{booktabs}      
\usepackage{multirow}      
\usepackage{amssymb}       

\usepackage{colortbl}
\usepackage{xcolor}

\usepackage{hyperref}
\usepackage{url}
\usepackage{multicol}
\usepackage{aliascnt}
\usepackage{adjustbox}
\usepackage{siunitx} 
\usepackage{booktabs}
\usepackage{multirow} 
\usepackage{enumitem}
\usepackage{booktabs} 
\usepackage{amssymb}  
\usepackage{amsmath}  
\usepackage{graphicx} 
\graphicspath{{./figs/}}
\usepackage{longtable} 
\usepackage{subcaption}
\usepackage{calc}
\usepackage[most]{tcolorbox}
\usepackage{listings}

\definecolor{uclablue}{rgb}{0.15, 0.45, 0.68}
\hypersetup{
    breaklinks,
    citecolor=uclablue,
    colorlinks=true,
}

\newtcolorbox{jianibox}[2][]{jianibox,title=#2,#1}
\tcbset{
  jianibox/.style={
    top=10pt,
    colframe=black,
    colbacktitle=black,
    coltitle=white,
    center,
  }
}

\lstdefinelanguage{prompt}{
    basicstyle=\scriptsize\ttfamily, 
    mathescape=true,        
    escapebegin=\color{latentcolor},  
    escapeend={},
    escapechar=@,
    stringstyle = \color{myorange},
    showstringspaces = false,
    moredelim = [s][\color{mypink}]{`}{`},
    moredelim = [s][\color{mybrown}]{```json}{```},
    moredelim = [s][\color{latentcolor}]{<StartOfLatent>}{<EndOfLatent>},
    literate = %
        {\ \ a.\ }{{\textcolor{mypurple}{\ \ a.\ }}}5
        {\ \ b.\ }{{\textcolor{mypurple}{\ \ b.\ }}}5
        {\ \ c.\ }{{\textcolor{mypurple}{\ \ c.\ }}}5
        {\ \ d.\ }{{\textcolor{mypurple}{\ \ d.\ }}}5
        {\ \ e.\ }{{\textcolor{mypurple}{\ \ e.\ }}}5
        {\ \ f.\ }{{\textcolor{mypurple}{\ \ f.\ }}}5
        {\ \ g.\ }{{\textcolor{mypurple}{\ \ g.\ }}}5
        {\ \ h.\ }{{\textcolor{mypurple}{\ \ h.\ }}}5
        {\ I.\ }{{\textcolor{mypurple}{\ I.\ }}}4
        {\ II.\ }{{\textcolor{mypurple}{\ II.\ }}}5
        {\ III.\ }{{\textcolor{mypurple}{\ III.\ }}}6
        {\ IV.\ }{{\textcolor{mypurple}{\ IV.\ }}}5
        {\ V.\ }{{\textcolor{mypurple}{\ V.\ }}}4
}

\newtcblisting[auto counter, number within=subsection]{prompt}[4][]{%
  before skip=6pt, after skip=6pt,    
  top=6pt, bottom=6pt, left=7pt, right=7pt, 
  enhanced, breakable,
  colback=white, colframe=gray!65,
  boxrule=0.6pt, arc=2pt,
  drop shadow={black!15},             
  listing only,
  listing options={
    language=prompt,
    upquote=true,
    basicstyle=\scriptsize\ttfamily \setlength{\baselineskip}{1.1\baselineskip},
    columns=fullflexible,
    breaklines=true,
    breakindent=0pt,
    xleftmargin=\ifx\empty#2\empty0pt\else#2\fi,
    xrightmargin=\ifx\empty#3\empty0pt\else#3\fi,
    aboveskip=0pt,                    
    belowskip=0pt,                    
    showstringspaces=false,
  },
  title={\ifstrempty{#4}{Prompt \thetcbcounter}{Prompt \thetcbcounter: #4}},
  colbacktitle=gray!6, coltitle=black,
  attach boxed title to top left={yshift=-1mm, xshift=6pt},
  boxed title style={colback=gray!6, boxrule=0pt, sharp corners},
  #1
}

\newtcblisting[auto counter, number within=subsection]{showcase}[4][]{%
  before=\par\vspace{\baselineskip},
  after=\par,
  width=\linewidth,
  enhanced,
  arc=0em,
  boxrule=1pt,
  listing only,
  listing options={
    language=prompt,
    upquote=true,
    basicstyle=\scriptsize\ttfamily \setlength{\baselineskip}{1.1\baselineskip},
    breaklines=true,
    breakindent=0pt,
    xleftmargin=\ifx\empty#2\empty-12pt\else#2\fi,
    xrightmargin=\ifx\empty#3\empty-5pt\else#3\fi,
    aboveskip=-4pt,
    belowskip=-4pt,
    columns=fullflexible,
    },
  colback=white,
  colframe=gray,
  colbacktitle=gray!5,
  coltitle=black,
  attach boxed title to top center={yshift=-3mm},
  box align=center,
  parbox=false,
  title={\ifstrempty{#4}{Example \thetcbcounter}{Example \thetcbcounter: #4}},
  #1
}

\definecolor{linkColor}{rgb}{0.2,0.4,0.6}
\definecolor{myblue}{HTML}{0379AC}
\definecolor{myred}{HTML}{A50E50}
\definecolor{myorange}{RGB}{238, 133, 74}
\definecolor{latentcolor}{named}{cyan}
\definecolor{normalcolor}{RGB}{0, 0, 0}
\usepackage{marvosym}

\usepackage{tabularx}
\usepackage{bbm}
\usepackage{makecell}

\usepackage{cleveref}
\newcommand{\minisection}[1]{\noindent{\textbf{#1}}.}

\usepackage{caption}
\title{More Than Seeing: Vision as Scalable Anchors in Native \\
Multimodal Pre-Training}

\author{
Haoyuan Wu$^{1, 2}$,
Aoqi Wu$^{2}$,
Jiani Hu$^{2}$,
Bei Yu$^{1}$
\\
\textbf{$^1$The Chinese University of Hong Kong} \quad 
\textbf{$^2$Foundation Model Department, Tencent}
}

\begin{document}
\maketitle

\begin{abstract}

Native multimodal pre-training trains a single model from scratch on text and multimodal data, enabling it to understand both language and visual inputs. 
However, how visual signals shape scaling behavior and support capabilities beyond visual understanding remains unclear. 
We investigate this question through controlled visual-content ablations across model sizes and training budgets. 
Specifically, we replace image embeddings with content-free padding token embeddings. 
Retaining visual content improves text prediction on image-text examples without materially changing the lowest text-only loss at a given compute budget, the optimal balance of model size and training data, or aggregate language scores.
The relative multimodal loss benefit grows approximately as a power law of compute.
To test benefits beyond understanding images, we evaluate spatial reasoning from text descriptions alone.
Both conditions improve with model and data scaling.
During training of the largest model, visual gains are most sustained in localization, navigation, and symbolic planning, which involve maintaining object configurations or composing spatial updates; gains in spatial relations, geometry, and physical commonsense are less consistent.
These findings identify visual content as a scalable anchor for multimodal learning, with benefits that extend beyond visual inputs while preserving aggregate language performance.

\end{abstract}

\section{Introduction}

Native multimodal pre-training develops visual and language capabilities in a single model trained from scratch on text and multimodal data~\citep{shukor2025scalingnmm,cui2025emu35,wu2026nmmscaling}.
Considering that images convey information about objects, scenes, and spatial relationships, their contribution may extend beyond tasks that require visual input.
This raises two related questions about how visual content affects scaling in native multimodal pre-training and whether it improves capabilities evaluated entirely through text.

Answering these questions requires isolating visual content from other changes introduced by multimodal training.
Scaling laws describe how training loss depends on model size, data volume, and training compute, guiding how to allocate training resources between parameters and tokens~\citep{shukor2025scalingnmm,wu2026nmmscaling}.
In native multimodal models, these relationships reflect learning from both text and multimodal data.
However, changing the multimodal data mixture can also alter language supervision and resource allocation, preventing direct attribution of differences in loss or downstream performance to visual content.
Isolating its contribution therefore requires varying visual content while holding language supervision and the training configuration constant.

We conduct this controlled comparison across model sizes and training compute budgets.
At each scale, we train matched models using either image patch embeddings or content-free padding, holding the architecture, training text, and training settings constant.
We analyze loss-compute frontiers and compute-optimal allocations separately for the text and multimodal objectives~\citep{wu2026nmmscaling}.
To quantify the contribution of visual content to multimodal understanding, we define the visual anchor gap as the relative increase in text prediction loss under the padding condition compared with the image condition.
We then examine how this gap scales with multimodal compute.

Our analyses reveal complementary scaling patterns.
Visual content leaves the text loss frontier and compute-optimal allocations nearly unchanged.
Meanwhile, it reduces multimodal loss, and the visual anchor gap follows an approximate power law of multimodal compute.
These findings suggest that the relative contribution of visual content to multimodal text prediction increases with compute, while pure-text scaling remains largely unchanged.

A growing visual anchor gap, however, does not establish a benefit when visual input is absent.
We therefore evaluate downstream capabilities using text-only inputs, both across model sizes and throughout training of the largest model.
General language benchmarks test whether the stable pure-text loss pattern is accompanied by comparable language performance.
Spatial tasks provide a targeted test of benefits from visual learning because images encode spatial relationships that can also be described and queried through language.

The image and padding conditions achieve similar aggregate performance on general language benchmarks across model sizes and training compute budgets.
Under both conditions, spatial reasoning performance improves with model size and additional training data.
During training of the largest model, visual gains are most sustained in localization, navigation, and symbolic planning, which involve maintaining object configurations or composing spatial updates.
Gains in spatial relations, geometry, and physical commonsense are less consistent, suggesting selective transfer rather than a uniform spatial advantage.
Long-plan success also emerges at larger scales, although execution constraints remain a bottleneck (\Cref{sec:spatial_emergence}).
These gains accompany similar aggregate general-language performance.
 Our main contributions can be summarized as follows:
\begin{itemize}[leftmargin=2em,topsep=0em]
\item We isolate the contribution of visual content through matched image and padding conditions across model sizes and training compute budgets, holding language supervision and the training configuration constant.
\item We identify closely matched pure-text loss frontiers and compute-optimal allocations, alongside a multimodal visual anchor gap that grows with compute under multiple allocation rules.
\item We examine capability development under text-only evaluation, showing preserved aggregate language performance and selective spatial transfer, with sustained gains in localization, navigation, and symbolic planning during A3B training.
\end{itemize}

\section{Preliminaries}
\label{sec:prelim}

\subsection{Native Multimodal Pre-Training}

Native multimodal pre-training trains a single model from scratch using interleaved multimodal data~\citep{shukor2025scalingnmm,cui2025emu35,wu2026nmmscaling}.
This approach enables both modalities to share parameter capacity and integrate fundamentally throughout training rather than being aligned post hoc. 
Specifically, images are mapped into continuous patch embeddings and interleaved with text tokens within a single decoder-only architecture. 
The model is trained autoregressively with a next-token prediction objective, with loss evaluated only at text tokens~\citep{shukor2025scalingnmm,wu2026nmmscaling}.
Under a fixed compute budget, native multimodal pre-training inherits the trade-off between model size and training data established for LLMs~\citep{kaplan2020oaiscaling,hoffmann2022chinchilla, wu2026nmmscaling}.
We use the compute-optimal framework and estimators developed for native multimodal pre-training~\citep{shukor2025scalingnmm,wu2026nmmscaling} to analyze the two streams under matched training conditions.

\subsection{Compute-Optimal Allocation}

Given a fixed compute budget $C$, we investigate the optimal allocation between model size $N$ and training token count $D$. 
Here $N$ counts activated non-embedding parameters per token, and $D$ counts all input tokens. 
Data composition is parameterized by $r=D_{\mathrm{mm}}/D_{\mathrm{text}}$, where multimodal tokens include both image-patch embeddings and text. 
Using the standard approximation $C=6ND$, compute accounting is
\begin{equation}
D=D_{\mathrm{text}}+D_{\mathrm{mm}}=(1+r)D_{\mathrm{text}},\;
C=6ND=C_{\mathrm{text}}+C_{\mathrm{mm}},\;
C_s=6ND_s,\; s\in\{\mathrm{text},\mathrm{mm}\}.
\label{eq:compute_accounting}
\end{equation}

Let $L_s^q(N,D_s;r)$ denote the smoothed training loss under condition $q\in\{\mathrm{img},\mathrm{pad}\}$. 
The two conditions retain or ablate visual content, respectively. 
Both source losses are evaluated at text tokens and optimized using shared parameters. 
When training examples are rarely revisited, smoothed training loss serves as a practical proxy for test loss~\citep{kaplan2020oaiscaling,hoffmann2022chinchilla}. 
We use it as the loss measure for our scaling analysis.
We estimate the preferred allocation separately for each objective using the same jointly trained models~\citep{wu2026nmmscaling}. 
At fixed $(s,q,r)$, the ideal allocation is
\begin{equation}
\bigl(N_s^{q,*}(C_s;r),D_s^{q,*}(C_s;r)\bigr)
\in\underset{N,D_s:\,6ND_s=C_s}{\arg\min}\;L_s^q(N,D_s;r).
\label{eq:alloc}
\end{equation}
Its loss defines the ideal frontier $L_s^{q,*}(C_s;r)$.
The definitions below allow general $r$. 
Our matched \textsc{img}/\textsc{pad} study fixes $r=0.3$, and we suppress this fixed argument in~\Cref{sec:method}.

\subsection{Estimators of the Compute-Optimal Frontier}

IsoFLOP profiles provide our primary allocation estimates while the training-curve envelope provides a sensitivity check using the same runs. 
Let $e\in\{\mathrm{iso},\mathrm{env}\}$ index the estimator. 
Hats denote finite-data estimates, in contrast to the ideal optima in~\Cref{eq:alloc}.
We summarize allocations by fitted power laws,
\begin{equation}
\widehat N_{s,e}^q(C_s;r)\propto C_s^{a_{s,e}^q(r)},\;
\widehat D_{s,e}^q(C_s;r)\propto C_s^{b_{s,e}^q(r)},\;
a_{s,e}^q(r)+b_{s,e}^q(r)=1,
\label{eq:powerlaw}
\end{equation}
The exponent constraint follows from $\widehat D=C_s/(6\widehat N)$ when the two log-log regressions use identical budgets and weights.
Budget selection and estimator details appear in~\Cref{appendix:preprocessing,appendix:allocation_estimators}.

\minisection{IsoFLOP profiles}
At a fixed budget $C_s$, each model size corresponds to $D_s=C_s/(6N)$. 
We read each model's loss in log compute, obtaining $\widetilde L_s^q(N,C_s;r)$, and fit a quadratic in $\log N$. 
A convex fit's minimum, clipped to the observed size range, estimates $\widehat N_{s,\mathrm{iso}}^q$. 
The token allocation follows algebraically. 
Regressing both allocations against compute in log space gives $a_{s,\mathrm{iso}}^q(r)$ and $b_{s,\mathrm{iso}}^q(r)$.

\minisection{Training-curve envelope}
Increasing $D_s$ at fixed $N$ traces a loss-compute trajectory. 
At each supported budget, the lowest interpolated loss across models defines the equal-FLOPs envelope $\widehat L_{s,\mathrm{env}}^q(C_s;r)$. 
Each envelope point retains its model size and source-token count. 
Regressing these selected allocations gives the second set of allocation exponents. 
Agreement between the estimators is a sensitivity check, not independent validation.

\minisection{The compute frontier}
We fit the equal-FLOPs envelope with
\begin{equation}
\mathcal F_s^q(C_s;r)
=E_s^q(r)+\left(\frac{C_{c,s}^q(r)}{C_s}\right)^{\beta_s^q(r)},\; 
C_{c,s}^q(r)>0,\quad\beta_s^q(r)>0.
\label{eq:power_law}
\end{equation}
Here $E_s^q(r)$ is a fitted loss offset, $C_{c,s}^q(r)$ a characteristic compute scale, and $\beta_s^q(r)$ a loss-decay exponent distinct from the token-allocation exponent $b_{s,e}^q(r)$.
The curve $\mathcal F_s^q$ describes the observed compute range and is distinct from the discrete envelope and ideal frontier. 
\Cref{appendix:frontier_fitting} gives the fitting procedure.

\subsection{Experimental Setup}
\minisection{Training data}
The nominal training corpus contains 250B text-source tokens from web pages, books, academic papers, and other domains. 
The multimodal corpus comprises web-crawled image-text pairs and interleaved image-text documents. 
With images represented as continuous patch embeddings, it contains 75B input tokens, including image positions and associated text. 
Thus, the full-training budget is $D=325\mathrm{B}$ with $r=0.3$; both conditions use the same source budgets.

\minisection{Model architecture}
We use a decoder-only MoE Transformer with a single patch-embedding layer that maps images directly into continuous embeddings without a pre-trained vision encoder. 
The models use auxiliary-loss-free MoE training~\citep{liu2024deepseekv3}. 
The scaling ladder contains six activated non-embedding parameter scales, $N\in\{71\mathrm{M},128\mathrm{M},340\mathrm{M},590\mathrm{M},874\mathrm{M},3\mathrm{B}\}$, denoted A71M through A3B.
Architecture and optimization details are provided in~\Cref{appendix:training}.
\section{Vision as a Scalable Anchor}
\label{sec:method}

Visual content can shape native multimodal learning in two ways.
It can provide context for image-text examples and influence shared model parameters that also support language learning.
Using a matched visual-content ablation across scales, we investigate whether vision changes the pure-text learning frontier.
We also examine whether its contribution to multimodal learning strengthens as compute increases.
The pure-text frontier remains nearly unchanged, whereas the relative loss advantage from visual content grows with multimodal compute.
We use visual anchoring to describe this contribution to multimodal understanding.

\subsection{Isolating the Contribution of Visual Content}

\minisection{A matched visual-content ablation}
Removing image-text examples would change both the visual inputs and the language supervision, making their effects difficult to separate.
Instead, we retain the examples and replace only their visual content.
At each of the six model sizes in~\Cref{sec:prelim}, we train a matched pair of models.
The \textsc{img} model receives image patch embeddings, whereas \textsc{pad} replaces each image embedding with a fixed, content-free pad-token embedding.
The two models share the same architecture, training text, data order, token budget, and optimization settings.
The replacement preserves sequence length, positional structure, and the text-loss mask.
Thus, the two conditions differ in access to visual content while retaining the same amount of multimodal data.

Both models are trained from scratch, with losses evaluated only on text tokens.
Their comparison therefore captures the combined effect of learning from visual content and using it as context during pre-training.
We analyze the text-only and multimodal streams separately, keeping the mixture fixed at $r=0.3$.
As illustrated in~\Cref{sec:prelim}, $C_s=6ND_s$ denotes compute attributed to stream $s$ within the jointly trained model.
Here, $D_{\mathrm{mm}}$ includes both image and text tokens.

\minisection{Measuring visual anchoring}
For stream $s\in\{\mathrm{text},\mathrm{mm}\}$, the anchor gap at allocation $(N,D_s)$ is the relative loss increase
\begin{equation}
\Delta_s^{\mathrm{anc}}(N,D_s)
=
\frac{L_{s}^{\mathrm{pad}}(N,D_s)-L_{s}^{\mathrm{img}}(N,D_s)}
{L_{s}^{\mathrm{img}}(N,D_s)}.
\label{eq:anchor_gap}
\end{equation}
A positive gap means that visual content lowers the loss.
A gap near zero means that the two conditions attain similar losses.
To compare allocations at fixed compute, we interpolate each condition's loss in log compute, following~\Cref{sec:prelim}, and obtain the paired relative gap
\begin{equation}
\widetilde\Delta_s^{\mathrm{anc}}(N,C_s)=
\frac{\widetilde L_s^{\mathrm{pad}}(N,C_s)-\widetilde L_s^{\mathrm{img}}(N,C_s)}{\widetilde L_s^{\mathrm{img}}(N,C_s)}.
\label{eq:interpolated_anchor_gap}
\end{equation}
We denote the model sizes supported by both conditions at budget $C_s$ by $\mathcal N_s(C_s)$.
Gaps are expressed as fractions in the equations and percentages in the figures.
This paired measure complements the frontier analysis by comparing the same allocation across conditions, whereas each condition’s frontier selects its own loss-minimizing allocation.

\begin{figure}[!tb]
\centering
\includegraphics[width=0.985\linewidth]{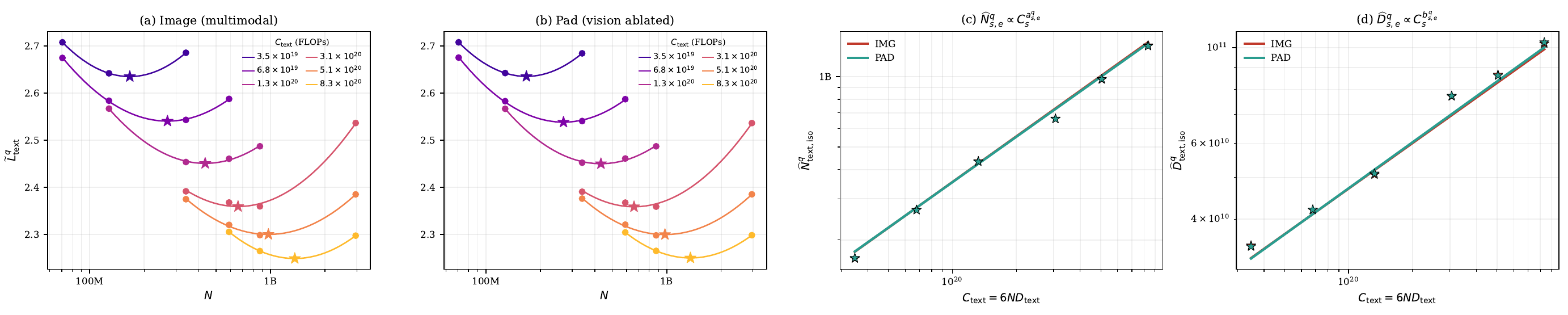}
\caption{\textbf{IsoFLOP curves (language objective). }
(a,b) At each target budget $C_{\mathrm{text}}$, we read the $\widetilde L_{\mathrm{text}}^{\mathrm{pad}}$ and $\widetilde L_{\mathrm{text}}^{\mathrm{img}}$ across various model size at the token count satisfying $6ND_{\mathrm{text}}=C_{\mathrm{text}}$ and plot it against $\log N$. (c,d) Optimal model size $\widehat N_{\mathrm{text},\mathrm{iso}}^q$ and optimal token count $\widehat D_{\mathrm{text},\mathrm{iso}}^q$ as a function of the compute budget $C_{\mathrm{text}}$.}
\label{fig:text_isoflop}
\end{figure}

\begin{figure}[!tb]
\centering
\includegraphics[width=0.985\linewidth]{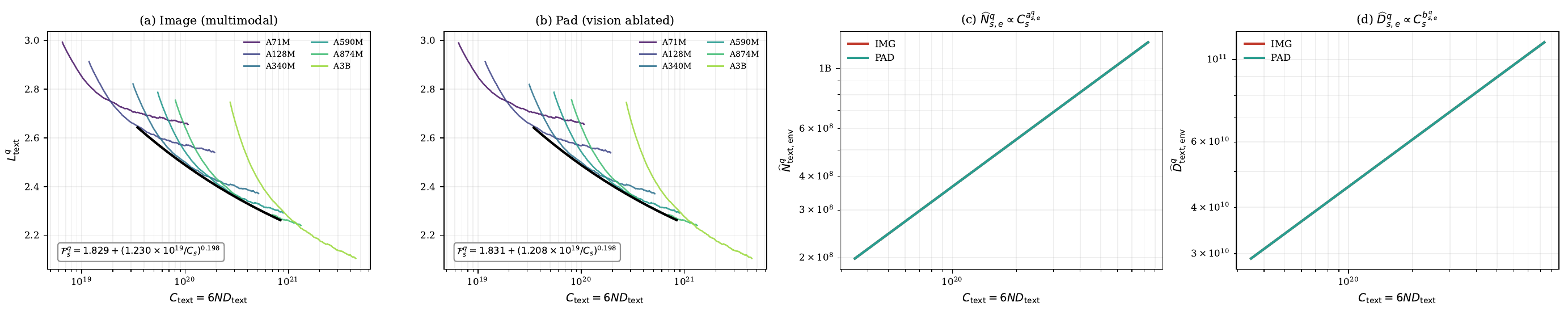}
\caption{\textbf{Training curve envelope (language objective).}
(a,b) Training curves span the evaluated model sizes.
The lower envelope gives the lowest observed loss at each budget $C_{\mathrm{text}}$. The fitted loss frontier summarizes this envelope. (c,d) Optimal model size $\widehat N_{\mathrm{text},\mathrm{env}}^q$ and optimal token count $\widehat D_{\mathrm{text},\mathrm{env}}^q$ as a function of the compute budget $C_{\mathrm{text}}$.}
\label{fig:text_envelope}
\end{figure}

\subsection{Visual Content Preserves Text Scaling}
\label{sec:text_invariance}

As shown in~\Cref{fig:text_isoflop}, the text IsoFLOP profiles closely align under \textsc{img} and \textsc{pad} at matched compute, including the locations of their minima.
The training-curve envelopes show the same pattern.
As illustrated in~\Cref{fig:text_envelope}, their fitted frontiers nearly overlap, with the same decay exponent $\beta_{\mathrm{text}}=0.198$ and closely matched offsets, $1.829$ and $1.831$.
Thus, visual-content ablation produces little change in the observed relationship between $L_{\mathrm{text}}^{q}$ and $C_{\mathrm{text}}$.
This stability extends to the estimated model-data allocation.
The IsoFLOP parameter exponents are $0.649$ for \textsc{img} and $0.645$ for \textsc{pad}.
The envelope estimator gives $0.578$ for both.
Although the absolute exponents depend on the estimator, the difference between conditions is small within each estimator.
Because both estimators reuse the same runs, we interpret this agreement as a sensitivity check on the allocation analysis.
The paired anchor gaps provide a direct check that this pattern is not solely a consequence of fitting frontiers.
As shown in~\Cref{fig:anchor_scale}a, $\widetilde\Delta_{\mathrm{text}}^{\mathrm{anc}}$ ranges from $-0.095\%$ to $+0.128\%$, with a mean of $-0.006\%$.
These near-zero differences support preservation of the measured text objective under the fixed mixture and compute budgets.
We examine whether this pattern extends to downstream language capabilities separately in~\Cref{sec:general_lang}.

\subsection{Visual Content Lowers Multimodal Loss}
\label{sec:mm_scaling}

\begin{figure}[!tb]
\centering
\includegraphics[width=0.985\linewidth]{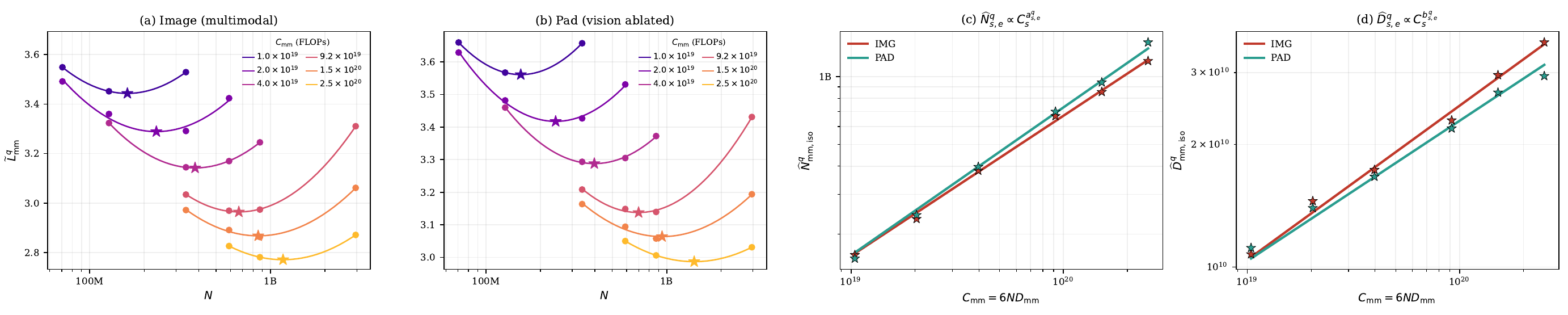}
\caption{\textbf{IsoFLOP curves (multimodal objective). }
(a,b) At each target budget $C_{\mathrm{mm}}$, we read the $\widetilde L_{\mathrm{mm}}^{\mathrm{pad}}$ and $\widetilde L_{\mathrm{mm}}^{\mathrm{img}}$ across various model size at the token count satisfying $6ND_{\mathrm{mm}}=C_{\mathrm{mm}}$ and plot it against $\log N$. (c,d) Optimal model size $\widehat N_{\mathrm{mm},\mathrm{iso}}^q$ and optimal token count $\widehat D_{\mathrm{mm},\mathrm{iso}}^q$ as a function of the compute budget $C_{\mathrm{mm}}$.}
\label{fig:mm_isoflop}
\end{figure}

\begin{figure}[!tb]
\centering
\includegraphics[width=0.985\linewidth]{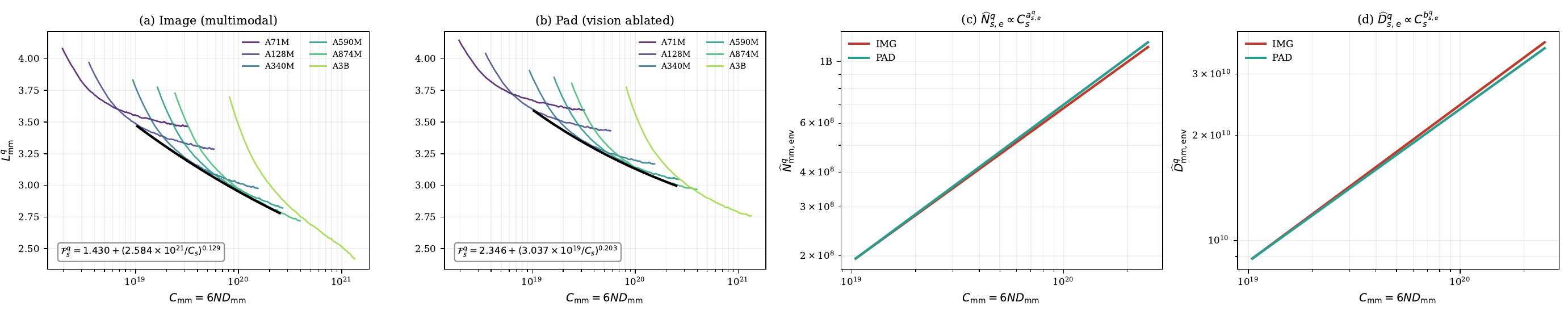}
\caption{\textbf{Training curve envelope (multimodal objective).}
(a,b) Training curves span the evaluated model sizes.
The lower envelope gives the lowest observed loss at each budget $C_{\mathrm{mm}}$. The fitted loss frontier summarizes this envelope. (c,d) Optimal model size $\widehat N_{\mathrm{mm},\mathrm{env}}^q$ and optimal token count $\widehat D_{\mathrm{mm},\mathrm{env}}^q$ as a function of the compute budget $C_{\mathrm{mm}}$.}
\label{fig:mm_envelope}
\end{figure}

As shown in~\Cref{fig:mm_isoflop}, matched \textsc{img} models attain lower loss than \textsc{pad} models on the text tokens in image-text examples at every evaluated budget.
As illustrated in~\Cref{fig:mm_envelope}, this advantage persists when each condition selects its own allocation.
The fitted \textsc{img} frontier lies below the \textsc{pad} frontier throughout $C_{\mathrm{mm}}\in[1.042\times10^{19},2.504\times10^{20}]$ FLOPs.
Visual content therefore lowers multimodal loss both at matched allocations and along the estimated loss frontiers.

\subsection{Visual Anchoring Strengthens with Compute}
\label{sec:anchor_scaling}

\begin{figure}[!tb]
\centering
\includegraphics[width=0.985\linewidth]{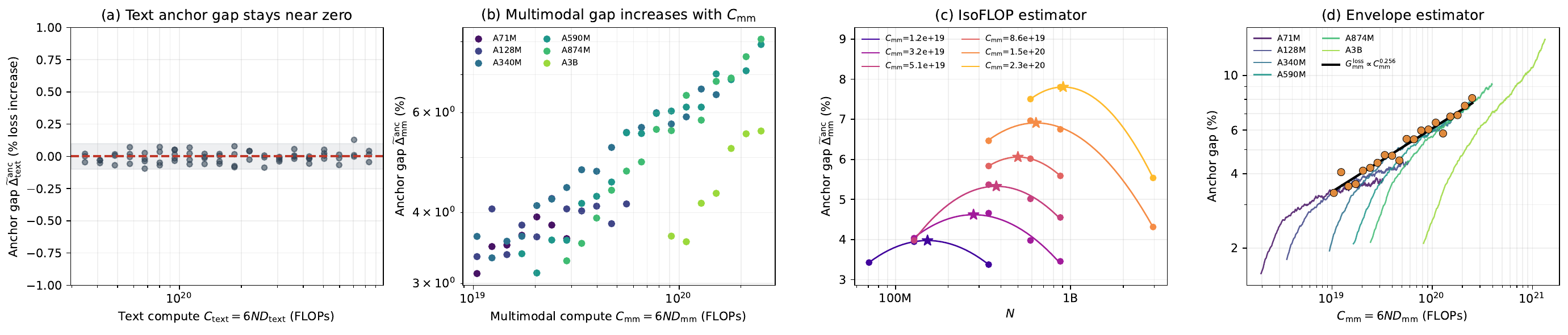}
\caption{\textbf{Visual anchoring strengthens with multimodal compute while the text gap stays near zero.}
(a) Paired text gaps, with a shaded $\pm0.1\%$ reference band. (b) Paired multimodal gaps, colored by model size. (c) Gap profiles at six fixed compute budgets. Stars mark maxima of quadratic fits in $\log N$, not loss minima. (d) Per-model gap trajectories, paired gaps along the \textsc{img} envelope, and their power-law fit (black, $\gamma=0.256$). All gaps are relative loss increases under ablation, expressed as percentages.}
\label{fig:anchor_scale}
\end{figure}

A lower multimodal frontier establishes that visual content helps, but it does not show whether its relative contribution persists as the model and data scale.
We therefore return to the anchor gap and track its dependence on multimodal compute.
As shown in~\Cref{fig:anchor_scale}, the gaps increase overall while fixed-compute profiles retain a dependence on model size.
To characterize the compute trend, we use three complementary summaries.

\minisection{Separating compute growth from allocation choice}
At each budget, $G_s^{\mathrm{max}}$ is the largest paired gap among the supported model sizes $\mathcal N_s(C_s)$.
A second summary, $G_s^{\mathrm{quad}}$, is the maximum of a quadratic fit to the paired gaps in $\log N$ over the supported size interval.
We find this maximum by evaluating both endpoints and any interior stationary point.
These summaries describe where the relative effect of ablation is largest.
To ask whether the gap also grows along an allocation chosen for low loss, we evaluate
\begin{equation}
G_s^{\mathrm{loss}}(C_s)
=\widetilde\Delta_s^{\mathrm{anc}}\!\left(\widehat N_{s,\mathrm{env}}^{\mathrm{img}}(C_s),C_s\right).
\label{eq:anchor_loss_path}
\end{equation}
Here, the \textsc{img} envelope selects the model size, and the gap compares \textsc{img} and \textsc{pad} at that same size and compute.
This is a paired gap along the \textsc{img} envelope, rather than a ratio between independently optimized frontiers.
Maximizing the gap and minimizing the loss need not select the same model.

\minisection{A consistent trend across allocation rules}
For each multimodal summary, we fit a power law by linear regression in log gap and log compute:
\begin{equation}
G_{\mathrm{mm}}^j(C_{\mathrm{mm}})\approx\kappa_{\mathrm{mm}}^j C_{\mathrm{mm}}^{\gamma_{\mathrm{mm}}^j},\;
j\in\{\mathrm{max},\mathrm{quad},\mathrm{loss}\},
\label{eq:anchor_C}
\end{equation}
with $C_{\mathrm{mm}}$ measured in FLOPs.
The three summaries yield closely matched exponents, $\gamma_{\mathrm{mm}}^{\mathrm{max}}=0.254$, $\gamma_{\mathrm{mm}}^{\mathrm{quad}}=0.254$, and $\gamma_{\mathrm{mm}}^{\mathrm{loss}}=0.256$, with log-space $R^2$ values of $0.978$, $0.972$, and $0.955$, respectively.
At $C_{\mathrm{mm}}=10^{20}$ FLOPs, the fitted gaps are $0.06231$, $0.06212$, and $0.06099$ in the same order.
From the smallest to the largest fitted budget, the quadratic maximum rises from $3.63\%$ to $8.09\%$, while the gap along the \textsc{img} envelope rises from $3.35\%$ to $8.09\%$.
This agreement shows that the overall increase is also present along allocations selected for low \textsc{img} loss, although local budget-to-budget fluctuations remain.

\minisection{Compute captures the trend, with allocation still mattering}
A descriptive joint fit gives $\widetilde\Delta_{\mathrm{mm}}^{\mathrm{anc}}\propto N^{0.212}D_{\mathrm{mm}}^{0.301}$.
The unequal exponents, together with the fixed-compute profiles in~\Cref{fig:anchor_scale}c, show that the gap is not determined by compute alone.
The power laws summarize an observed scaling trend across the tested configurations.
Because the budgets and summaries reuse correlated observations from the same runs, the quality of these fits does not establish training-seed reproducibility or behavior beyond the fitted range.
We provide more details in~\Cref{appendix:method_estimation}.
\begin{figure}[!tb]
\centering
\includegraphics[width=0.56\linewidth]{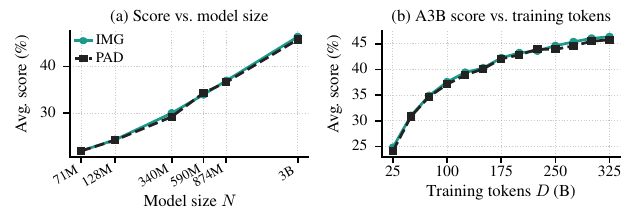}
\caption{\textbf{Aggregate language performance.} 
Average scores over 15 text benchmarks for \textsc{img} and \textsc{pad}
(a) Text performance across model sizes $N$. (b) A3B text performance over $D$.}
\label{fig:language_preserved}
\end{figure}

\begin{figure}[!tb]
\centering
\includegraphics[width=0.99\linewidth]{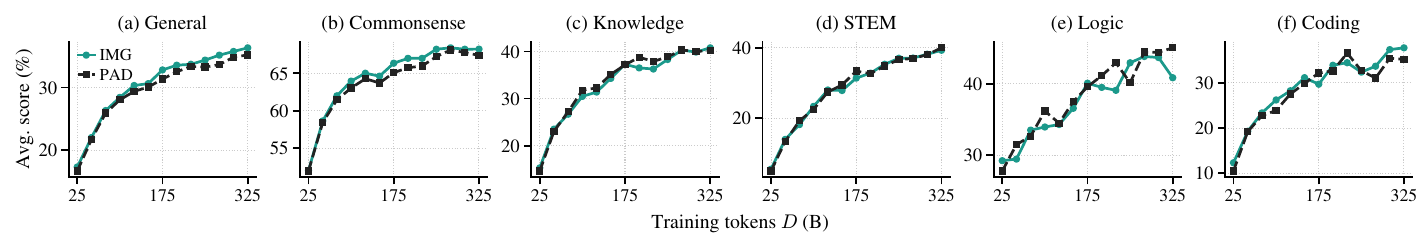}
\caption{\textbf{Per-category language performance.} Average scores evaluated using A3B models for \textsc{img} and \textsc{pad} conditions.}
\label{fig:language_category_trajectories}
\end{figure}

\section{General Language Capability}
\label{sec:general_lang}

The text scaling law in~\Cref{sec:text_invariance} remains nearly unchanged when visual content is ablated.
We further investigate whether the contribution of vision to multimodal learning is accompanied by a change in general language performance. 
Using the same matched \textsc{img} and \textsc{pad} runs, we compare mean scores over 15 text benchmarks spanning general understanding, commonsense, knowledge, mathematics, logical reasoning, and coding. 
We provide more details in \Cref{appendix:eval_general}.

\minisection{Aggregate language performance follows a shared scaling trend}
As shown in~\Cref{fig:language_preserved}, increasing model size produces substantial gains under both conditions, while the \textsc{img} and \textsc{pad} curves remain closely aligned. 
The difference between conditions stays small relative to the improvement across the scaling ladder. 
\Cref{fig:language_preserved} shows the same pattern along the data axis.
Specifically, both conditions improve rapidly early in training and more gradually thereafter, with closely matched aggregate scores.
The stable pure-text loss pattern is therefore accompanied by preserved aggregate language performance in the evaluated suite. 

\minisection{Category gains accompany matched aggregate performance}
As shown in~\Cref{fig:language_category_trajectories}, \textsc{img} achieves higher mean scores in coding, general understanding, and commonsense reasoning than \textsc{pad}. 
These gains accompany closely matched aggregate language performance shown in~\Cref{fig:language_preserved}. 

\section{Spatial Reasoning}
\label{sec:spatial}

Visual anchoring in~\Cref{sec:method} concerns multimodal text prediction. 
We now investigate spatial reasoning about object locations, relations, and configurations under text-only evaluation.
Using the same matched \textsc{img} and \textsc{pad} runs, we evaluate 15 benchmarks across six task groups, averaging benchmarks equally within each group and then weighting groups equally.
Evaluation settings and full scores appear in~\Cref{appendix:eval_spatial,table:spatial_img,table:spatial_pad,table:pt_spatial_img,table:pt_spatial_pad}.

\subsection{Spatial Reasoning Scales with Compute}
\label{sec:spatial_scaling}

\begin{figure}[!tb]
\centering
\includegraphics[width=0.56\linewidth]{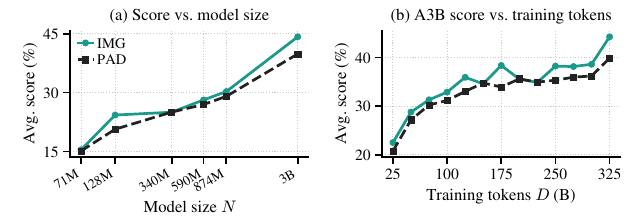}
\caption{\textbf{Aggregate spatial reasoning performance.}
Scores for \textsc{img} and \textsc{pad}, averaged equally across six task groups, with equal benchmark weights within each group.
(a) Final performance across model sizes $N$ at a nominal budget of $D=325\mathrm{B}$ tokens. (b) A3B performance over nominal training tokens $D$.}
\label{fig:spatial_scale}
\end{figure}

At the nominal budget $D=325\mathrm{B}$, aggregate accuracy improves at every successive model size under both conditions.
From A71M to A3B, \textsc{img} rises from 15.41\% to 44.25\% and \textsc{pad} from 15.10\% to 39.87\% (\Cref{fig:spatial_scale}). 
At fixed A3B size, increasing $D$ from 25B to 325B raises accuracy from 22.56\% to 44.25\% for \textsc{img} and from 20.81\% to 39.87\% for \textsc{pad}, with local fluctuations and a further gain at the final checkpoint. 
Thus, substantial spatial capability develops through scaling even without visual content.

\subsection{Visual Content Adds a Measurable Gain}
\label{sec:spatial_transfer}

Visual content adds a smaller, scale-dependent benefit. 
The matched \textsc{img}-\textsc{pad} gap is positive at five of six model sizes and reaches 4.38\% at A3B. 
It is not monotonic.
Specifically, the gap is 3.62\% at A128M and approximately 1.2\% at A590M and A874M. 
These comparisons distinguish overall capability growth from the additional contribution of visual pre-training.

\subsection{Transfer Varies Across Task Groups}
\label{sec:spatial_selective}

\begin{figure}[!tb]
\centering
\includegraphics[width=0.99\linewidth]{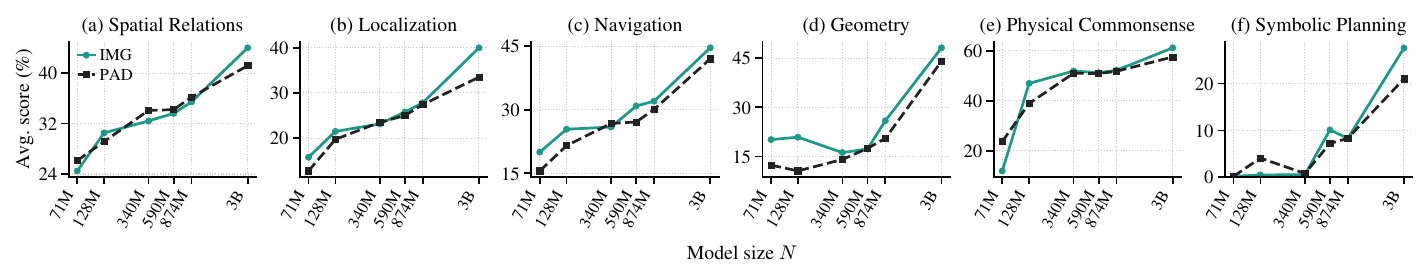}
\caption{\textbf{Scores across model sizes for each task group.}}
\label{fig:spatial_family_scale}
\end{figure}

\begin{figure}[!tb]
\centering
\includegraphics[width=0.99\linewidth]{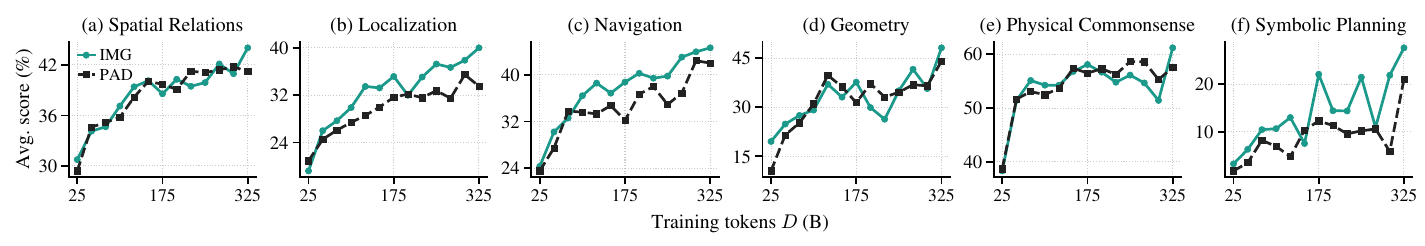}
\caption{\textbf{Scores under data scaling for each task group.}}
\label{fig:spatial_skills}
\end{figure}

Every task group improves from A71M to A3B under both conditions, but visual gains vary across groups and scales (\Cref{fig:spatial_family_scale}).
During A3B training, localization, navigation, and symbolic planning have mean \textsc{img}-\textsc{pad} gains of 2.88\%, 3.02\%, and 5.19\%, respectively (\Cref{fig:spatial_skills}).
The other three groups have mean gaps within 0.6\% of zero, although all six favor \textsc{img} at the final checkpoint.
The distinction is thus the persistence of gains across training, not their presence at a single checkpoint.

The three groups with sustained gains share demands for maintaining or composing spatial configurations: localization binds objects to positions, navigation updates positions along routes, and symbolic planning composes actions toward a target arrangement.
The other groups emphasize relation inference, geometric rules, or prior physical knowledge, offering solution routes through language and symbolic constraints.
PIQA-spatial, for example, scales strongly with model size despite little average visual advantage during A3B training; geometry gains concentrate in shape identification, while Geometry3K remains difficult.
Together, these findings suggest more consistent transfer to organizing configurations and spatial updates.
These demands overlap, so this interpretation does not establish a mechanism or strict taxonomy (\Cref{appendix:spatial_analysis}).

\subsection{Emergence of Compositional Spatial Skills}
\label{sec:spatial_emergence}

Symbolic planning has the largest mean \textsc{img}-\textsc{pad} gain during A3B training, approximately 5.19\%.
Blocksworld accuracy rises from 3.3\% to 27.6\% for \textsc{img}, versus 1.8\% to 21.0\% for \textsc{pad} (\Cref{fig:spatial_skills}).
Success requires a complete executable action sequence that reaches the goal; any invalid action or unmet goal invalidates the response.

\begin{figure}[!tb]
\centering
\includegraphics[width=0.56\linewidth]{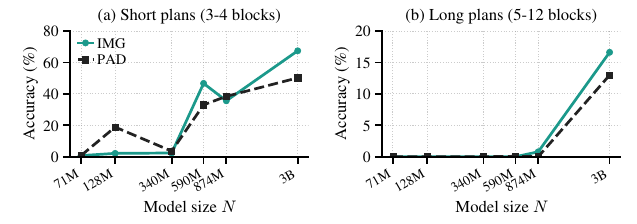}
\caption{Growth of symbolic tower-building performance in the benchmark Blocksworld. Final-checkpoint scores for (a) short plans with 3-4 blocks (106 problems) and (b) long plans with 5-12 blocks (385 problems) in the tower-building task.}
\label{fig:spatial_planning}
\end{figure}

As shown in~\Cref{fig:spatial_planning}, larger models extend successful composition to longer plans.
Here, emergence denotes the appearance of successful long-plan composition at larger scales, without establishing a sharp scaling transition or the acquisition of individual atomic operations.
On 3-4-block problems, \textsc{img} reaches 46.70\% at A590M and 67.45\% at A3B. On 5-12-block problems, no \textsc{img} response succeeds through A590M; success rises to 0.78\% at A874M and 16.62\% at A3B, compared with 12.99\% for A3B \textsc{pad}. 
\Cref{appendix:spatial_analysis} gives the response counts, and the ten-block case in~\Cref{sec:spatial_showcase} illustrates a complete \textsc{img} solution versus a legal but incomplete \textsc{pad} prefix.

\begin{figure}[!tb]
\centering
\includegraphics[width=\linewidth]{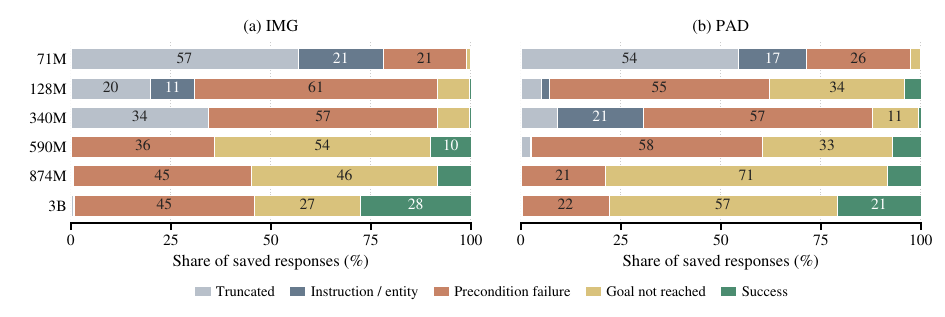}
\caption{Execution outcomes across scales for (a) \textsc{img} and (b) \textsc{pad}. Length-limited responses are classified first; others are executed until the first invalid instruction or unmet precondition, with a goal check after legal plans. Labels give percentages for segments of at least 10\%.}
\label{fig:spatial_plan_errors}
\end{figure}

Execution remains a bottleneck. Our final-checkpoint audit distinguishes truncation, instruction/entity errors, precondition violations, and legal plans that miss the goal (\Cref{appendix:spatial_analysis,fig:spatial_plan_errors}). 
Truncation dominates A71M but becomes uncommon at A3B.
At A3B, \textsc{img} produces fewer legal-but-unsuccessful plans than \textsc{pad} (523 versus 1,121), but more precondition failures (882 versus 427). 
Among legal-but-unsuccessful plans, 73.4\% under \textsc{pad} contain fewer actions than minimally required, compared with 44.2\% under \textsc{img}.

This contrast is consistent with better organization of target configurations or action sequences, as suggested in~\Cref{sec:spatial_selective}, while reliable execution remains a bottleneck.
Many \textsc{img} precondition failures attempt to pick up a covered block: placing $a$ on $b$ too early prevents subsequently placing $b$ on $c$.
The text-prediction objective does not explicitly enforce these constraints.
Competing first-failure categories do not isolate improvements in individual operations; \Cref{appendix:spatial_analysis,sec:spatial_showcase} provide the full audit and cases.

\section{Conclusion}
Our controlled visual-content ablations show that visual information acts as a scalable anchor in native multimodal pre-training. 
Its contribution to multimodal text prediction grows with compute, while pure-text loss scaling, compute-optimal allocation, and aggregate general-language performance remain largely unchanged. 
Spatial reasoning improves with model and data scaling under both conditions, and visual content provides an additional benefit even when evaluation inputs contain only text. 
During A3B training, gains are most sustained in localization, navigation, and symbolic planning, which share demands for maintaining configurations or composing spatial updates.
Gains in spatial relations, geometry, and physical commonsense, with greater emphasis on explicit relations, rules, or prior knowledge, are less consistent.
This suggests selective spatial transfer; long-plan success emerges at larger scales, while execution remains a bottleneck.

\clearpage
{
\bibliographystyle{iclr2025_conference}
\bibliography{ref/Top, ref/reference, ref/spatial_references}
}

\appendix
\section{Related Work}
\label{sec:related_work}

\minisection{Native multimodal pre-training}
Multimodal systems differ in when and how vision and language enter a shared learning process.
Modular approaches connect a pre-trained vision encoder to a language model through bridging modules or instruction tuning~\citep{li2023blip2,liu2023llava}.
Native approaches instead train on mixed-modal sequences within a shared backbone. 
Chameleon uses discrete image and text tokens~\citep{chameleon2024}, Transfusion combines language modeling with image diffusion~\citep{zhou2024transfusion}, and Emu3.5 studies unified autoregressive modeling of large-scale interleaved vision-language data~\citep{cui2025emu35}.
In this study, we investigate how the information carried by these inputs affects learning across scales.

\minisection{Scaling laws and cross-modal interactions}
Language-model scaling laws relate loss to model size, data, and compute and formalize the allocation of a fixed training budget~\citep{kaplan2020oaiscaling,hoffmann2022chinchilla}.
Mixed-modal scaling extends this analysis to interactions between data sources, including synergy and competition~\citep{aghajanyan2023mixedmodal}.
Subsequent work examines native multimodal architectures and sparsity~\citep{shukor2025scalingnmm}, data-mixture weights~\citep{shukor2026datamixscaling}, and differences between language and vision scaling requirements~\citep{tong2026beyondlang}.
Our prior study estimates allocations for the text and multimodal objectives separately and analyzes their dependence on the training mixture~\citep{wu2026nmmscaling}.
Changing the mixture, however, can also alter language supervision. Tong et al. investigate caption-distribution shifts as a possible source of language degradation~\citep{tong2026beyondlang}.
We hold the text, mixture, sequence structure, and optimization settings fixed while ablating visual content. 
This comparison targets image information within a common training setup.

\minisection{Visual context and transfer to language}
Visual ablation can probe a trained model's use of image context~\citep{frank2021crossmodal}.
VisNec uses losses with and without visual context to select informative instruction-tuning examples~\citep{dong2026visnec}.
Its blind forward pass changes the input to a single model, whereas our anchor gap compares matched models trained with different access to visual content.
A complementary line of work studies how visual supervision affects language learning through vokenization~\citep{tan2020vokenization}, visual-knowledge probing~\citep{alper2023bertblind}, and word learning under limited text data~\citep{zhuang2024grounding}.
Work on visual-document pre-training also compares continued learning from scientific pages with learning from parsed text of the same documents~\citep{zhang2026visualpretraining}.
These settings motivate examining both the contribution of visual context to training loss and the capabilities retained when visual inputs are absent. We study both within native pre-training while preserving the associated text across conditions.

\minisection{Spatial reasoning without visual inputs}
Textual spatial benchmarks assess relations and operations that can also be expressed visually.
StepGame evaluates relation composition and multi-hop reasoning~\citep{hjn_shi2022_stepgame}, while SpatialEval compares textual and visual input conditions~\citep{wang2024spatialeval}.
Other methods introduce visual computation during inference~\citep{menon2024whiteboard,li2025mvot}. 
Our evaluation instead uses text-only inputs and outputs after pre-training.
Our prior work reports spatial improvements when we add multimodal data to a fixed text budget~\citep{wu2026nmmscaling}. 
The present matched ablation retains both the associated text and the input budget.
This design separates capability growth with scale from the additional contribution of visual content, while leaving the internal mechanism of this transfer unresolved.

\section{Training settings}
\label{appendix:training}
For native multimodal pre-training, we employ the Muon optimizer with a weight decay of 0.1 and a gradient clipping threshold of 1.0. 
Additionally, we utilize a warmup-stable learning rate schedule, a global batch size of 2048, and a maximum sequence length of 4096 (8M input tokens per step).
The training loss is computed only on text tokens, with vision tokens masked out from loss computation~\citep{kimi2026kimi25}.
We provide more details in~\Cref{table:arch}.

\begin{table}[!tb]
\centering
\setlength{\tabcolsep}{8pt}
\scriptsize
\caption{Pre-training hyperparameters.
We detail the hyperparameters used for pre-training different model configurations to derive scaling laws.}
\begin{tabular}{lcccccc}
\toprule
& \textbf{A71M} & \textbf{A128M} & \textbf{A340M} & \textbf{A590M} & \textbf{A874M} & \textbf{A3B} \\
\midrule
\multicolumn{7}{l}{\textbf{Architecture Configurations}} \\
\midrule
layers & 11 & 15 & 19 & 23 & 27 & 48 \\
hidden size & 640 & 768 & 1152 & 1280 & 1536 & 2048 \\
FFN hidden size & 2048 & 2048 & 3072 & 4096 & 4096 & 6912 \\
attention heads & 8 & 12 & 16 & 20 & 24 & 32 \\
query groups & \multicolumn{6}{c}{4} \\
kv channels & \multicolumn{6}{c}{128} \\
total experts & \multicolumn{6}{c}{128} \\
activated experts & \multicolumn{6}{c}{8} \\
expert FFN hidden size & 256 & 256 & 384 & 512 & 512 & 768 \\
shared-expert hidden size & 256 & 256 & 384 & 512 & 512 & 768 \\
MoE layer & \multicolumn{6}{c}{first layer dense, remaining layers MoE} \\
\midrule
\multicolumn{7}{l}{\textbf{Training Settings}} \\
\midrule
batch size (sequences) & \multicolumn{6}{c}{2048} \\
sequence length & \multicolumn{6}{c}{4096} \\
warmup steps & \multicolumn{6}{c}{2000} \\
learning rate & 2.6e-3 & 1.9e-3 & 1.1e-3 & 8.2e-4 & 6.8e-4 & 3.0e-4 \\
\bottomrule
\end{tabular}
\label{table:arch}
\end{table}

\section{Evaluation settings}

\subsection{General language capability}
\label{appendix:eval_general}

We evaluate general language capabilities using 15 distinct benchmarks.
\begin{itemize}[topsep=0pt, leftmargin=2em]
    \item \textbf{Aggregate:} SuperGPQA\citep{du2025supergpqa}(5-shot), MMLU-Redux\citep{hendrycks2020mmlu}(5-shot), MMLU-Pro\citep{wang2024mmlupro}(5-shot), and AGIEval$_{\mathrm{en}}$\citep{zhong2023agieval}(3-shot).
    \item \textbf{Coding:} HumanEval+\citep{liu2023evalplus}(0-shot) and MBPP+\citep{liu2023evalplus}(3-shot).
    \item \textbf{Mathematics:} GSM8K\citep{cobbe2021gsm8k}(4-shot, CoT) and MATH\citep{lightman2023math500}(4-shot, CoT).
    \item \textbf{Logical Reasoning:} BBH\citep{suzgun2022bbh}(3-shot, CoT).
    \item \textbf{Knowledge:} NaturalQuestions\citep{kwiatkowski2019naturalquestions}~(5-shot) and TriviaQA\citep{joshi2017triviaqa}(5-shot).
    \item \textbf{Commonsense Reasoning:} HellaSwag\citep{zellers2019hellaswag}(10-shot), WinoGrande\citep{sakaguchi2021winogrande}(5-shot), SIQA\citep{sap2019socialiqa}(0-shot), and PIQA\citep{bisk2020piqa}(0-shot).
\end{itemize}
For the evaluation of general language capability, we use greedy generation strategy.

\subsection{Spatial reasoning}
\label{appendix:eval_spatial}
We evaluate 15 benchmarks in six task groups, all with 5-shot chain-of-thought prompting and text-only inputs.
Each entry below combines the group's scope with its benchmarks.
\begin{itemize}[topsep=0pt, leftmargin=2em]
    \item \textbf{Spatial Relations:} Infer or verify relative positions.
    StepGame~\citep{hjn_shi2022_stepgame} uses directional statements; SpaRTUN~\citep{mirzaee-kordjamshidi-2022-transfer} covers direction, distance, containment, and contact; bAbI(positional reasoning)~\citep{hjn_weston2015_babi} tests positional constraints; and FoREST~\citep{hjn_premsri2025_forest} varies the reference frame.
    \item \textbf{Localization:} Identify, describe, or count objects using their attributes and positions.
    SpartQA~\citep{mirzaee-etal-2021-spartqa}, SpatialMap~\citep{wang2024spatialeval}, and BBH(colored objects)~\citep{suzgun2022bbh} query objects within described scenes.
    \item \textbf{Navigation:} Follow routes, update positions or orientations, or construct paths.
    SpatialUnderstanding~\citep{hjn_yamada2024_spatialeval} queries the object reached after successive moves; MazeNav~\citep{wang2024spatialeval} uses text-represented mazes; bAbI(path finding)~\citep{hjn_weston2015_babi} requires a direction sequence; and BBH(navigate)~\citep{suzgun2022bbh} tests return to the starting point.
    \item \textbf{Geometry:} Recover geometric configurations and apply their constraints.
    BBH(geometric shapes)~\citep{suzgun2022bbh} identifies shapes from SVG path coordinates; Geometry3K~\citep{hjn_lu2021_intergps} uses textual logic forms for lengths, angles, and areas, without diagram images.
    \item \textbf{Physical Commonsense:} Choose between solutions to everyday physical goals.
    PIQA-spatial~\citep{bisk2020piqa} tests knowledge of object use, materials, or support, rather than general physical simulation.
    \item \textbf{Symbolic Planning:} Produce a complete executable action sequence satisfying all goal relations. 
    Blocksworld~\citep{valmeekam2023planbench} asks the model to build a tower from separate blocks, tracking which blocks are clear, held, or placed, and respecting action preconditions.
\end{itemize}
We generate four responses per item with temperature 0.3 and top-$p=0.9$, reporting avg@4.
Each benchmark belongs to one group; benchmarks are weighted equally within groups, and the six group scores are averaged equally.
These groups organize this evaluation rather than define a general taxonomy of spatial or physical reasoning.
\Cref{tab:spatial_inventory} lists item counts and answer types.

\begin{table}[!tb]
\centering
\scriptsize
\setlength{\tabcolsep}{6.4pt}
\caption{Spatial evaluation inventory. The 15 benchmarks cover six task groups.}
\label{tab:spatial_inventory}
\begin{tabular}{@{}llrl@{}}
\toprule
Capability & Benchmark & Items & Answer type \\
\midrule
\multirow{4}{*}{Spatial Relations} & StepGame & 3,000 & Direction label \\
 & SpaRTUN & 3326 & Relation label\\
 & bAbI(positional reasoning) & 627 & Yes/no \\
 & FoREST & 1998 & Direction label \\
\midrule
\multirow{3}{*}{Localization} & SpartQA & 1994 & Choice label \\
 & SpatialMap & 997 & Choice label \\
 & BBH(colored objects) & 245 & Choice label \\
\midrule
\multirow{4}{*}{Navigation} & SpatialUnderstanding & 3998 & Object name \\
 & MazeNav & 999 & Choice label \\
 & bAbI(path finding) & 995 & Direction sequence \\
 & BBH(navigate) & 245 & Yes/no \\
\midrule
\multirow{2}{*}{Geometry} & BBH(geometric shapes) & 244 & Choice label \\
 & Geometry3K & 596 & Choice label \\
\midrule
Physical Commonsense & PIQA-spatial & 548 & Choice label \\
\midrule
Symbolic Planning & Blocksworld & 491 & Action sequence \\
\bottomrule
\end{tabular}
\end{table}

\section{Allocation Estimation and Loss-Compute Frontiers}
\label{appendix:method_estimation}
This section details the estimators in~\Cref{sec:prelim,sec:method}, retaining the source $s\in\{\mathrm{text},\mathrm{mm}\}$, condition $q\in\{\mathrm{img},\mathrm{pad}\}$, and estimator $e\in\{\mathrm{iso},\mathrm{env}\}$ indices. 
We suppress the fixed mixture argument $r$ and report results at $r=0.3$. 
The compute is $C_s=6ND_s$ in FLOPs. 
Hats denote finite-data estimates, tildes denote fixed-compute interpolated losses, and $\mathcal F_s^q$ denotes a fitted frontier. 
The loss-decay exponent $\beta_s^q$, allocation exponents $a_{s,e}^q,b_{s,e}^q$, and gap-growth exponent $\gamma_s^j$ have distinct meanings.

\subsection{Data Preprocessing}
\label{appendix:preprocessing}
With $T=2048\times4096 \approx 8\mathrm{M}$ input tokens per optimizer step, step $k$ corresponds to
\begin{equation}
D=Tk,\;
D_{\mathrm{text}}=\frac{D}{1+r},\;
D_{\mathrm{mm}}=\frac{rD}{1+r},\; C_s=6ND_s.
\label{eq:step_token_accounting}
\end{equation}
Here $D$ is the total input-token count and $D_s$ is the token count for source $s$. 
The 325B total-token budget is nominal, whereas the loss-compute analysis uses actual selected steps.
For each source and condition at fixed $r$, we construct various candidate budgets uniformly in log compute over the union of the model-specific ranges. 
Writing $C_{s,m}^{q,\min}$ and $C_{s,m}^{q,\max}$ for model $m$'s endpoints, we define the supported model set as
\begin{equation}
\mathcal M_s^q(C_s)=
\{m:C_{s,m}^{q,\min}\leq C_s\leq C_{s,m}^{q,\max}\}.
\label{eq:model_support}
\end{equation}
We retain budgets with $|\mathcal M_s^q(C_s)|\geq3$ and interpolate losses linearly in $\log C_s$ within each model's observed range, without extrapolation, to obtain $\widetilde L_s^q(N_m,C_s)$. 

\subsection{Estimating Compute-Optimal Allocation}
\label{appendix:allocation_estimators}
\minisection{IsoFLOP profiles}
At each retained budget and fixed $(s,q)$, let $x=\log N$ and fit the losses with a quadratic. 
For this source, condition, and budget, we estimate coefficients $p_0,p_1,p_2$ in the quadratic
\begin{equation}
P_{s,C_s}^q(x)=p_2x^2+p_1x+p_0.
\label{eq:isoflop_quadratic}
\end{equation}
For $p_2>0$, the allocation estimate is
\begin{equation}
\log\widehat N_{s,\mathrm{iso}}^q(C_s)
=\operatorname{clip}\!\left(-\frac{p_1}{2p_2},x_{\min},x_{\max}\right),\;
\widehat D_{s,\mathrm{iso}}^q(C_s)=\frac{C_s}{6\widehat N_{s,\mathrm{iso}}^q(C_s)},
\label{eq:isoflop_estimate}
\end{equation}
where $x_{\min},x_{\max}$ delimit the model sizes supported at that budget. 
We regress $\log\widehat N_{s,\mathrm{iso}}^q$ and $\log\widehat D_{s,\mathrm{iso}}^q$ on $\log C_s$ with equal weights. 
This yields $a_{s,\mathrm{iso}}^q$ and $b_{s,\mathrm{iso}}^q$, with $a_{s,\mathrm{iso}}^q+b_{s,\mathrm{iso}}^q=1$ by the compute identity.

\minisection{Training-curve envelope}
The equal-FLOPs envelope is
\begin{equation}
\widehat L_{s,\mathrm{env}}^q(C_s)
:=\min_{m\in\mathcal M_s^q(C_s)}\widetilde L_s^q(N_m,C_s).
\label{eq:appendix_envelope}
\end{equation}
Each point retains the minimizing model size $\widehat N_{s,\mathrm{env}}^q(C_s)$ and token count $\widehat D_{s,\mathrm{env}}^q(C_s)=C_s/(6\widehat N_{s,\mathrm{env}}^q(C_s))$. 
Regressing these allocations in log space gives $a_{s,\mathrm{env}}^q$ and $b_{s,\mathrm{env}}^q$, again summing to one. 
Fitting the envelope losses gives the distinct frontier parameters below. 
Both allocation estimators reuse the same runs, so their comparison in~\Cref{fig:text_isoflop,fig:text_envelope,fig:mm_isoflop,fig:mm_envelope} is a sensitivity check rather than independent validation.

\subsection{Fitting Loss-Compute Frontiers}
\label{appendix:frontier_fitting}
At fixed $(s,q,r)$, \Cref{eq:power_law} can be written
\begin{equation}
\mathcal F_s^q(C_s)=E_s^q+\left(\frac{C_{c,s}^q}{C_s}\right)^{\beta_s^q}
=E_s^q+K_s^q C_s^{-\beta_s^q},\qquad K_s^q=(C_{c,s}^q)^{\beta_s^q}.
\label{eq:frontier_amplitude}
\end{equation}
The amplitude $K_s^q$ is distinct from the characteristic compute scale $C_{c,s}^q$. 
We search various equally spaced values of $\beta_s^q$ in $[0.01,0.60]$. 
For each candidate, ordinary least squares in loss space estimates $(E_s^q,K_s^q)$ from the various envelope points. 
Candidates with $K_s^q\leq0$ or any nonpositive prediction at a fitted budget are rejected. Among the remaining candidates, we minimize the mean squared log-loss residual and recover $C_{c,s}^q=(K_s^q)^{1/\beta_s^q}$. 
This two-stage selection is not joint least squares in log loss; it does not constrain $E_s^q>0$, although all four reproduced fits have positive offsets. 
Each source and condition is fitted independently at $r=0.3$.

\subsection{Fitting Visual Anchoring}
The anchor analysis applies~\Cref{eq:interpolated_anchor_gap} to paired losses. 
We read each condition's loss at the same model size and compute budget before taking their relative difference.
For the quadratic maximum, we fit paired gaps in centered $\log N$ and check both endpoints and any interior stationary point. 
The loss-path summary uses the discrete model selected by the \textsc{img} envelope at each budget, with the \textsc{pad} loss evaluated at that same allocation. 
It is not a ratio of independently optimized or fitted frontiers.

Positive multimodal summaries are regressed in log gap against log compute. Figures express gaps as percentages, whereas~\Cref{eq:anchor_C} defines them as fractions. If the percentage regression has intercept $h_j$, then
\begin{equation}
\log\!\left(100G_{\mathrm{mm}}^j(C_{\mathrm{mm}})\right)
\approx h_j+\gamma_{\mathrm{mm}}^j\log C_{\mathrm{mm}},\qquad
\kappa_{\mathrm{mm}}^j=\frac{\exp(h_j)}{100}.
\label{eq:gap_amplitude}
\end{equation}
Thus $\kappa_{\mathrm{mm}}^j$ is the coefficient of $C_{\mathrm{mm}}^{\gamma_{\mathrm{mm}}^j}$ when compute is expressed in FLOPs. 
For $j=\mathrm{max},\mathrm{quad},\mathrm{loss}$, respectively, it is approximately $5.1874\times10^{-7}$, $5.1693\times10^{-7}$, and $4.6420\times10^{-7}$. 
The gaps reported at $10^{20}$ FLOPs in~\Cref{sec:method} are fitted function values, not these coefficients.

\section{General Language Evaluation Results}
\label{appendix:exp}

We report the benchmark-level scores underlying the general-language comparisons in~\Cref{sec:general_lang}.

\section{Spatial Evaluation Results}
\label{appendix:spatial_protocol}

We report the benchmark-level scores underlying the spatial comparisons in~\Cref{sec:spatial}.

\subsection{Spatial Transfer and Planning Analysis}
\label{appendix:spatial_analysis}

\minisection{Benchmark-level interpretation}
Beyond the selective gains in~\Cref{sec:spatial_selective}, performance across model sizes is comparable in the remaining groups, whose averages mask distinct benchmark behaviors. 
Physical commonsense has the largest end-to-end improvement with model size, but a much smaller visual-content gain at A3B. 
Its benchmark, PIQA-spatial, selects between two solutions to an everyday physical goal and can draw on physical commonsense and language knowledge; scaling alone therefore does not establish a specifically visual mechanism. 
Geometry gains concentrate in BBH(geometric shapes), while Geometry3K remains substantially harder. 
These differences should be considered when interpreting the group averages reported alongside the full benchmark scores in~\Cref{table:spatial_img,table:spatial_pad,table:pt_spatial_img,table:pt_spatial_pad}.

\minisection{Plan Length and Execution Outcomes}
Blocksworld requires an executable action sequence satisfying all goal relations, unlike answer selection in PIQA-spatial. 
This strict criterion helps explain why visual-content gains coexist with limited absolute success. 
The long-plan results in~\Cref{fig:spatial_planning} use 1540 responses to 385 problems with 5-12 blocks.
No \textsc{img} response succeeds through A590M, while 12 succeed at A874M (0.78\%) and 256 at A3B (16.62\%), compared with 200 for A3B \textsc{pad} (12.99\%). 
The short-plan results are reported in~\Cref{sec:spatial_emergence}.

For the execution analysis in~\Cref{fig:spatial_plan_errors}, each response receives one outcome. 
Truncation at the length limit takes precedence; otherwise, execution stops at the first uninterpretable instruction, undeclared or unspecified block, or violated action precondition (e.g., stacking a block that is not held). 
Goal attainment is checked only when every action is legal. 
These competing first-failure categories do not isolate improvements in individual operations.

\begin{table}[!tb]
\centering
\small
\setlength{\tabcolsep}{6pt}
\caption{Supplementary execution statistics. The last row is conditional on legal plans that miss the goal; its denominator differs from the outcome counts above.}
\label{tab:spatial_execution_summary}
\begin{tabular}{@{}llrr@{}}
\toprule
Scale & Outcome or subset & \textsc{img} & \textsc{pad} \\
\midrule
A71M & Truncation & 1,115 & 1,069 \\
A3B & Truncation & 14 & 0 \\
A3B & Legal plan, goal missed & 523 & 1,121 \\
A3B & Precondition violation & 882 & 427 \\
A3B & Too short among legal goal misses & 231/523 (44.2\%) & 823/1,121 (73.4\%) \\
\bottomrule
\end{tabular}
\end{table}

\Cref{tab:spatial_execution_summary} shows that truncation becomes uncommon at A3B, although its decline is not monotonic across intermediate sizes. 
Repetitive action loops in A71M outputs suggest that improved repetition and termination control may contribute. 
At A3B, \textsc{img} has approximately 53\% fewer legal-but-unsuccessful plans but 2.1 times as many precondition failures as \textsc{pad}; its unsuccessful legal plans are also less often shorter than minimally required. 
Many precondition failures attempt to pick up a covered block, sometimes immediately after placing another block on it. 
This contrast motivates the distinction between representing target configurations and executing plans in~\Cref{sec:spatial_emergence}; the ten-block example in~\Cref{sec:spatial_showcase} illustrates the distinction without establishing its prevalence.

\subsection{Showcases: What the Model Actually Does}
\label{sec:spatial_showcase}

\minisection{A complete long plan instead of a plausible prefix}
The ten-block example below makes the composition gain concrete. 
Each consecutive pickup-and-stack pair is abbreviated as $x\!\to\!y$, meaning \emph{pick up $x$; stack $x$ on $y$}. Initial states and goals were recovered from the public source instances; all 491 local reference plans execute successfully against their matched instances.

\begin{jianibox}[colback=white,colframe=hjn!55!black,colbacktitle=hjn!7,coltitle=hjn,coltext=hjn,top=5pt,bottom=5pt,fontupper=\small]{{Case 1: Completing a ten-block tower}}
\textbf{Problem (condensed).} Blocks $a,c,d,e,f,g,h,i,j,k$ are clear and on the table; the hand is empty. Build the tower $i$--$d$--$c$--$e$--$g$--$h$--$j$--$k$--$a$--$f$ (top to bottom).\par\smallskip
\textbf{A590M IMG, first response:}\quad $a\to f,\ d\to a,\ c\to d,\ e\to c$.\par
All eight actions are legal, but only one of nine target relations is met.\par\smallskip
\textbf{A3B PAD, first response:}\quad $a\to f,\ i\to a,\ d\to i$.\par
All six actions are legal, but only one of nine target relations is met.\par\smallskip
\textbf{A3B IMG, first response:}\quad $a\to f,\ k\to a,\ j\to k,\ h\to j,\ g\to h,\ e\to g,\ c\to e,\ d\to c,\ i\to d$.\par
All 18 actions are legal, and all nine target relations are met.
\end{jianibox}

A3B \textsc{img} succeeds on all four responses to this item, while every smaller \textsc{img} model and A3B \textsc{pad} fail on all four. The successful plan builds from the bottom upward, preserving earlier placements. Both unsuccessful prefixes shown above terminate normally: their failure concerns goal completion, rather than the output limit.

\minisection{A correct answer can conceal a broken spatial derivation}
The next example asks for the destination after moving right, down, right, up, left from white-headed capuchin.

\begin{jianibox}[colback=white,colframe=hjn!55!black,colbacktitle=hjn!7,coltitle=hjn,coltext=hjn,top=5pt,bottom=5pt,fontupper=\small]{{Case 2: Accepted answer, inconsistent route (\texttt{square-3431})}}
\textbf{Correct route} ($x$ rightward, $y$ downward): white-headed capuchin $(0,0)$ $\to$ lacewing $(1,0)$ $\to$ wallaby $(1,1)$ $\to$ hare $(2,1)$ $\to$ cassette $(2,0)$ $\to$ lacewing $(1,0)$.\par\smallskip
\textbf{A3B IMG explanation (condensed):} replaces the down move with another right move, assigning wallaby to $(2,0)$, then hare to $(2,1)$ and cassette to $(3,1)$; its final coordinate-to-object assignment is also inconsistent.\par
\textbf{Final answers:} IMG: \texttt{lacewing} (accepted); PAD: \texttt{hare} (incorrect).
\end{jianibox}

Multiple movement and entity errors make the written derivation invalid despite the accepted answer. They cannot be repaired by correcting one entity name, but also do not establish that the answer was guessed randomly.

\minisection{Separating verbal slips from failed operations}
A590M \textsc{img}'s first response executes the correct plan $f\to b,\ j\to f,\ g\to j$, despite once writing \emph{a clear on top} where the block is $f$. This local prose slip leaves the solution intact. By contrast, the following reversal changes the actual path:

\begin{jianibox}[colback=white,colframe=hjn!55!black,colbacktitle=hjn!7,coltitle=hjn,coltext=hjn,top=5pt,bottom=5pt,fontupper=\small]{{Case 3: Reversing only one of two relations (bAbI task 19, item 0)}}
\textbf{Relevant facts (condensed):} the bathroom is north of the bedroom; the bedroom is north of the hallway. Find a path from the bathroom to the hallway.\par
\textbf{Correct path:} south, south (\texttt{s,s}).\par
\textbf{A3B IMG, first response:} south, north (\texttt{s,n}).
\end{jianibox}

Only the first north-of relation is inverted correctly; executing the predicted path returns to the bathroom. Checking every edge in the travel direction would expose the error. These cases illustrate mechanisms; the exhaustive counts in \Cref{fig:spatial_plan_errors} remain specific to Blocksworld.

\minisection{Counting explanation quality within a fixed case panel}
We revisit eight fixed problems across all 12 conditions, inspecting each first response (96 traces). Of 27 accepted answers, 13 have consistent derivations, two contain local verbal errors, nine contain substantive errors, and three remain borderline (\Cref{tab:spatial_trace_quality}). The panel covers StepGame, three SpatialUnderstanding layouts, bAbI pathfinding, and Blocksworld. It supports same-problem case comparisons, but neither population frequency estimates nor a monotonic scaling trend in explanation reliability.

\begin{table}[!tb]
\centering\footnotesize
\setlength{\tabcolsep}{8pt}
\begin{tabular}{lccccc}
\toprule
& \multicolumn{2}{c}{IMG} && \multicolumn{2}{c}{PAD} \\
\cmidrule{2-3}\cmidrule{5-6}
Scale & Accepted & C / L / S / U && Accepted & C / L / S / U \\
\midrule
A71M  & 3 & 0 / 0 / 3 / 0 && 1 & 1 / 0 / 0 / 0 \\
A128M & 1 & 1 / 0 / 0 / 0 && 3 & 1 / 0 / 2 / 0 \\
A340M & 3 & 1 / 0 / 1 / 1 && 2 & 0 / 0 / 1 / 1 \\
A590M & 2 & 1 / 0 / 0 / 1 && 2 & 1 / 1 / 0 / 0 \\
A874M & 1 & 1 / 0 / 0 / 0 && 3 & 2 / 0 / 1 / 0 \\
A3B   & 4 & 2 / 1 / 1 / 0 && 2 & 2 / 0 / 0 / 0 \\
\bottomrule
\end{tabular}
\caption{Explanation quality among accepted answers in eight fixed problems per condition. 
C: consistent derivation; L: local verbal error with the solution intact; S: substantive reasoning error; U: unresolved boundary between local and substantive error. 
The four counts sum to Accepted, which is out of eight first responses. 
This manual case audit is separate from the all-response planning analysis and does not estimate population frequencies.}
\label{tab:spatial_trace_quality}
\end{table}

The analyses connect capability growth to both scaling axes, selective visual-content gains, and longer successful action compositions. 
These results concern single training runs under the archived evaluator. 
Executable plans and explanation checks complement final-answer accuracy.

\clearpage

\begin{table}[!tb]\centering\scriptsize
\setlength\tabcolsep{4.2pt}
\caption{\textbf{A3B \textsc{IMG} language scores across various $D$.}}
\label{table:pt_text_mm75}
\begin{tabular}{l|l|ccccccccccccc}
\toprule
 & \# Pre-Trained Tokens & 25B & 50B & 75B & 100B & 125B & 150B & 175B & 200B & 225B & 250B & 275B & 300B & 325B \\
\midrule
\multirow{4}{*}{\makecell[l]{Aggregate}} & MMLU-Redux (Acc.) & 27.05 & 38.81 & 44.74 & 48.35 & 51.19 & 51.37 & 54.49 & 54.21 & 55.05 & 56.00 & 57.17 & 57.09 & 57.98 \\
 & MMLU-Pro (Acc.) & 10.85 & 12.99 & 17.73 & 19.19 & 19.86 & 20.21 & 22.03 & 23.84 & 24.61 & 24.92 & 25.33 & 27.53 & 27.68 \\
 & AGIEval$_{\rm en}$ (Acc.) & 20.49 & 22.71 & 27.88 & 29.27 & 32.13 & 32.66 & 36.03 & 36.58 & 35.02 & 35.99 & 37.10 & 36.92 & 37.73 \\
 & SuperGPQA (Acc.) & 10.78 & 13.48 & 14.92 & 16.88 & 18.11 & 18.27 & 18.69 & 19.63 & 20.17 & 20.63 & 21.08 & 21.53 & 22.00 \\
\midrule
\multirow{2}{*}{\makecell[l]{Coding}} & HumanEval+ (Pass@1) & 6.10 & 11.59 & 15.24 & 14.63 & 15.24 & 15.24 & 13.41 & 18.90 & 19.51 & 15.85 & 17.68 & 21.34 & 21.34 \\
 & MBPP+ (Pass@1) & 18.52 & 26.98 & 31.48 & 37.83 & 41.27 & 47.09 & 46.03 & 48.94 & 49.47 & 48.94 & 49.73 & 53.44 & 54.23 \\
\midrule
\multirow{2}{*}{\makecell[l]{Mathematics}} & GSM8K (Pass@1) & 4.85 & 17.21 & 22.21 & 30.63 & 37.00 & 36.01 & 42.15 & 43.82 & 48.07 & 50.19 & 49.58 & 51.40 & 52.39 \\
 & MATH (Pass@1) & 5.95 & 10.60 & 14.05 & 16.05 & 19.00 & 19.70 & 20.65 & 21.80 & 22.60 & 23.90 & 24.65 & 25.00 & 26.30 \\
\midrule
\multirow{1}{*}{\makecell[l]{Logic Reasoning}} & BBH (EM) & 29.21 & 29.44 & 33.50 & 33.93 & 34.28 & 36.56 & 40.07 & 39.49 & 39.08 & 42.94 & 43.82 & 43.68 & 40.84 \\
\midrule
\multirow{2}{*}{\makecell[l]{Knowledge}} & NaturalQuestions (EM) & 10.08 & 14.99 & 16.32 & 18.17 & 19.31 & 21.77 & 24.13 & 22.05 & 19.64 & 22.02 & 23.35 & 21.55 & 23.63 \\
 & TriviaQA (EM) & 20.56 & 32.08 & 37.08 & 42.78 & 43.47 & 46.81 & 50.42 & 50.97 & 52.92 & 54.58 & 57.22 & 58.19 & 57.92 \\
\midrule
\multirow{4}{*}{\makecell[l]{Commonsense \\ Reasoning}} & HellaSwag (Acc.) & 51.23 & 59.40 & 63.20 & 66.47 & 66.97 & 67.87 & 68.27 & 69.37 & 69.53 & 70.60 & 71.23 & 71.37 & 71.57 \\
 & SIQA (Acc.) & 34.85 & 45.14 & 52.71 & 54.15 & 56.24 & 55.48 & 59.72 & 59.67 & 60.90 & 62.28 & 62.08 & 61.87 & 62.95 \\
 & PIQA (Acc.) & 71.16 & 74.59 & 75.63 & 76.11 & 77.31 & 76.88 & 77.42 & 77.31 & 78.29 & 77.86 & 79.05 & 78.45 & 77.80 \\
 & WinoGrande (Acc.) & 50.83 & 55.41 & 56.43 & 59.12 & 59.51 & 58.25 & 59.98 & 61.64 & 59.27 & 62.12 & 61.25 & 61.17 & 60.62 \\
\midrule
\rowcolor{blue!5} \multicolumn{2}{c|}{Average} & 24.83 & 31.03 & 34.88 & 37.57 & 39.39 & 40.28 & 42.23 & 43.22 & 43.61 & 44.59 & 45.36 & 46.04 & 46.33 \\
\bottomrule\end{tabular}%
\end{table}

\begin{table}[!tb]\centering\scriptsize
\setlength\tabcolsep{6pt}
\caption{\textbf{\textsc{img} language scores across various $N$.}}
\label{table:text_mm75}
\begin{tabular}{l|l|cccccc}
\toprule
 & Model Sizes ($N$) & A71M & A128M & A340M & A590M & A874M & A3B \\
\midrule
\multirow{4}{*}{\makecell[l]{Aggregate}} & MMLU-Redux (Acc.) & 25.40 & 32.10 & 38.91 & 44.95 & 49.03 & 57.98 \\
 & MMLU-Pro (Acc.) & 10.68 & 11.63 & 13.87 & 14.97 & 17.29 & 27.68 \\
 & AGIEval$_{\rm en}$ (Acc.) & 20.64 & 20.45 & 25.54 & 26.29 & 29.67 & 37.73 \\
 & SuperGPQA (Acc.) & 8.00 & 10.31 & 11.81 & 14.16 & 14.90 & 22.00 \\
\midrule
\multirow{2}{*}{\makecell[l]{Coding}} & HumanEval+ (Pass@1) & 4.88 & 7.93 & 12.20 & 16.46 & 15.85 & 21.34 \\
 & MBPP+ (Pass@1) & 7.94 & 13.23 & 26.45 & 30.95 & 37.57 & 54.23 \\
\midrule
\multirow{2}{*}{\makecell[l]{Mathematics}} & GSM8K (Pass@1) & 2.27 & 5.08 & 14.25 & 23.27 & 28.05 & 52.39 \\
 & MATH (Pass@1) & 3.70 & 4.95 & 9.40 & 11.95 & 15.60 & 26.30 \\
\midrule
\multirow{1}{*}{\makecell[l]{Logic Reasoning}} & BBH (EM) & 26.87 & 28.06 & 30.31 & 33.03 & 35.94 & 40.84 \\
\midrule
\multirow{2}{*}{\makecell[l]{Knowledge}} & NaturalQuestions (EM) & 7.15 & 10.03 & 13.85 & 14.65 & 18.86 & 23.63 \\
 & TriviaQA (EM) & 17.08 & 17.92 & 25.97 & 36.94 & 39.86 & 57.92 \\
\midrule
\multirow{4}{*}{\makecell[l]{Commonsense \\ Reasoning}} & HellaSwag (Acc.) & 42.13 & 48.13 & 57.57 & 62.10 & 65.60 & 71.57 \\
 & SIQA (Acc.) & 32.45 & 33.88 & 43.04 & 49.90 & 52.05 & 62.95 \\
 & PIQA (Acc.) & 68.50 & 69.64 & 73.45 & 74.05 & 75.68 & 77.80 \\
 & WinoGrande (Acc.) & 51.54 & 51.70 & 53.83 & 56.35 & 58.41 & 60.62 \\
\midrule
\rowcolor{blue!5} \multicolumn{2}{c|}{Average} & 21.95 & 24.34 & 30.03 & 34.00 & 36.96 & 46.33 \\
\bottomrule\end{tabular}%
\end{table}

\clearpage

\begin{table}[!tb]\centering\scriptsize
\setlength\tabcolsep{4.2pt}
\caption{\textbf{A3B \textsc{pad} language scores across various $D$.}}
\label{table:pt_text_pad_mm75}
\begin{tabular}{l|l|ccccccccccccc}
\toprule
 & \# Pre-Trained Tokens & 25B & 50B & 75B & 100B & 125B & 150B & 175B & 200B & 225B & 250B & 275B & 300B & 325B \\
\midrule
\multirow{4}{*}{\makecell[l]{Aggregate}} & MMLU-Redux (Acc.) & 25.89 & 38.30 & 44.72 & 48.49 & 50.44 & 51.63 & 52.74 & 54.63 & 54.40 & 55.74 & 56.72 & 56.32 & 57.75 \\
 & MMLU-Pro (Acc.) & 11.17 & 12.20 & 15.96 & 16.94 & 18.53 & 18.59 & 20.05 & 21.95 & 21.62 & 22.29 & 22.20 & 24.36 & 25.49 \\
 & AGIEval$_{\rm en}$ (Acc.) & 19.16 & 23.64 & 28.36 & 31.37 & 33.51 & 33.49 & 34.94 & 35.63 & 37.57 & 35.60 & 36.13 & 37.86 & 37.42 \\
 & SuperGPQA (Acc.) & 10.43 & 12.63 & 14.66 & 15.33 & 15.04 & 16.57 & 17.68 & 17.84 & 19.89 & 19.44 & 19.61 & 20.87 & 20.17 \\
\midrule
\multirow{2}{*}{\makecell[l]{Coding}} & HumanEval+ (Pass@1) & 6.71 & 10.37 & 9.76 & 13.41 & 18.29 & 14.02 & 18.90 & 20.12 & 19.51 & 12.80 & 15.85 & 16.46 & 18.29 \\
 & MBPP+ (Pass@1) & 14.29 & 28.04 & 35.71 & 34.39 & 36.77 & 45.77 & 45.50 & 45.24 & 53.70 & 52.65 & 46.30 & 54.50 & 52.12 \\
\midrule
\multirow{2}{*}{\makecell[l]{Mathematics}} & GSM8K (Pass@1) & 4.40 & 16.07 & 23.50 & 29.49 & 38.59 & 39.50 & 45.19 & 43.44 & 46.70 & 50.72 & 50.04 & 49.81 & 53.83 \\
 & MATH (Pass@1) & 5.40 & 10.50 & 14.90 & 15.50 & 16.30 & 19.40 & 21.60 & 21.90 & 23.00 & 22.65 & 23.95 & 26.35 & 26.25 \\
\midrule
\multirow{1}{*}{\makecell[l]{Logic Reasoning}} & BBH (EM) & 27.77 & 31.47 & 32.64 & 36.17 & 34.47 & 37.50 & 39.58 & 41.12 & 42.94 & 40.14 & 44.44 & 44.40 & 45.04 \\
\midrule
\multirow{2}{*}{\makecell[l]{Knowledge}} & NaturalQuestions (EM) & 8.50 & 15.12 & 17.40 & 19.28 & 21.88 & 22.22 & 22.66 & 24.35 & 22.71 & 23.43 & 24.77 & 24.18 & 23.68 \\
 & TriviaQA (EM) & 20.83 & 30.97 & 37.08 & 44.31 & 42.64 & 48.06 & 51.67 & 53.19 & 53.06 & 54.31 & 55.83 & 55.83 & 56.67 \\
\midrule
\multirow{4}{*}{\makecell[l]{Commonsense \\ Reasoning}} & HellaSwag (Acc.) & 50.90 & 59.70 & 64.03 & 66.00 & 67.20 & 67.90 & 68.47 & 70.07 & 69.83 & 70.50 & 71.13 & 71.60 & 71.20 \\
 & SIQA (Acc.) & 33.47 & 44.37 & 48.82 & 51.38 & 52.30 & 50.97 & 53.94 & 54.04 & 55.53 & 58.65 & 59.52 & 58.91 & 58.65 \\
 & PIQA (Acc.) & 71.06 & 73.56 & 75.90 & 76.17 & 78.02 & 77.15 & 77.09 & 78.07 & 77.53 & 76.88 & 78.89 & 77.42 & 78.62 \\
 & WinoGrande (Acc.) & 52.41 & 56.04 & 57.06 & 58.56 & 59.59 & 58.80 & 60.85 & 60.77 & 61.01 & 62.90 & 62.90 & 63.22 & 61.09 \\
\midrule
\rowcolor{blue!5} \multicolumn{2}{c|}{Average} & 24.16 & 30.87 & 34.70 & 37.12 & 38.91 & 40.10 & 42.06 & 42.82 & 43.93 & 43.91 & 44.55 & 45.47 & 45.75 \\
\bottomrule\end{tabular}%
\end{table}

\begin{table}[!tb]\centering\scriptsize
\setlength\tabcolsep{6pt}
\caption{\textbf{\textsc{pad} language scores across various $N$.}}
\label{table:text_pad_mm75}
\begin{tabular}{l|l|cccccc}
\toprule
 & Model Sizes ($N$) & A71M & A128M & A340M & A590M & A874M & A3B \\
\midrule
\multirow{4}{*}{\makecell[l]{Aggregate}} & MMLU-Redux (Acc.) & 25.44 & 30.68 & 38.51 & 45.79 & 47.60 & 57.75 \\
 & MMLU-Pro (Acc.) & 11.18 & 11.63 & 13.22 & 16.48 & 16.70 & 25.49 \\
 & AGIEval$_{\rm en}$ (Acc.) & 21.42 & 21.73 & 24.94 & 27.60 & 30.52 & 37.42 \\
 & SuperGPQA (Acc.) & 8.39 & 9.90 & 12.65 & 13.55 & 13.83 & 20.17 \\
\midrule
\multirow{2}{*}{\makecell[l]{Coding}} & HumanEval+ (Pass@1) & 7.32 & 7.93 & 10.98 & 16.46 & 13.41 & 18.29 \\
 & MBPP+ (Pass@1) & 8.99 & 12.17 & 22.75 & 31.48 & 35.98 & 52.12 \\
\midrule
\multirow{2}{*}{\makecell[l]{Mathematics}} & GSM8K (Pass@1) & 1.82 & 5.16 & 13.34 & 22.52 & 27.37 & 53.83 \\
 & MATH (Pass@1) & 3.60 & 4.20 & 6.65 & 12.80 & 16.90 & 26.25 \\
\midrule
\multirow{1}{*}{\makecell[l]{Logic Reasoning}} & BBH (EM) & 25.55 & 28.09 & 30.68 & 31.66 & 35.26 & 45.04 \\
\midrule
\multirow{2}{*}{\makecell[l]{Knowledge}} & NaturalQuestions (EM) & 7.76 & 9.06 & 11.55 & 16.76 & 18.23 & 23.68 \\
 & TriviaQA (EM) & 14.72 & 18.75 & 26.39 & 35.00 & 39.31 & 56.67 \\
\midrule
\multirow{4}{*}{\makecell[l]{Commonsense \\ Reasoning}} & HellaSwag (Acc.) & 41.87 & 47.43 & 56.40 & 61.83 & 65.00 & 71.20 \\
 & SIQA (Acc.) & 32.34 & 35.82 & 42.78 & 51.33 & 54.81 & 58.65 \\
 & PIQA (Acc.) & 68.28 & 69.04 & 73.18 & 74.43 & 75.52 & 78.62 \\
 & WinoGrande (Acc.) & 50.75 & 52.17 & 54.77 & 57.70 & 59.20 & 61.09 \\
\midrule
\rowcolor{blue!5} \multicolumn{2}{c|}{Average} & 21.96 & 24.25 & 29.25 & 34.36 & 36.64 & 45.75 \\
\bottomrule\end{tabular}%
\end{table}

\clearpage

\begin{table}[!tb]\centering\scriptsize
\setlength\tabcolsep{3.6pt}
\caption{\textbf{A3B \textsc{img} spatial scores across various $D$.}}
\label{table:pt_spatial_img}
\begin{tabular}{l|l|ccccccccccccc}
\toprule
 & \# Pre-Trained Tokens & 25B & 50B & 75B & 100B & 125B & 150B & 175B & 200B & 225B & 250B & 275B & 300B & 325B \\
\midrule
\multirow{4}{*}{Spatial Relations} & StepGame & 16.34 & 19.31 & 21.07 & 24.62 & 27.58 & 26.99 & 27.02 & 29.59 & 29.45 & 29.04 & 29.68 & 31.56 & 33.11 \\
 & SpaRTUN & 19.97 & 26.72 & 24.69 & 29.19 & 27.83 & 31.30 & 27.44 & 30.37 & 27.64 & 32.22 & 32.34 & 32.25 & 32.17 \\
 & bAbI(positional reasoning) & 47.49 & 47.85 & 47.05 & 48.52 & 52.07 & 54.19 & 53.79 & 54.27 & 51.20 & 52.31 & 59.41 & 51.36 & 58.41 \\
 & FoREST & 39.18 & 42.62 & 45.78 & 46.11 & 50.10 & 48.01 & 46.03 & 46.91 & 49.54 & 46.01 & 47.00 & 48.65 & 52.41 \\
\midrule
\multirow{3}{*}{Localization} & SpartQA & 22.93 & 33.09 & 32.35 & 31.86 & 31.81 & 33.55 & 29.61 & 29.41 & 27.62 & 33.38 & 31.01 & 33.35 & 31.78 \\
 & SpatialMap & 20.91 & 26.45 & 29.01 & 28.91 & 33.45 & 33.30 & 33.27 & 33.10 & 31.47 & 34.25 & 35.21 & 32.22 & 33.25 \\
 & BBH(colored objects) & 13.67 & 18.37 & 21.63 & 28.88 & 35.10 & 32.76 & 42.55 & 33.37 & 46.02 & 44.08 & 43.78 & 48.06 & 54.90 \\
\midrule
\multirow{4}{*}{Navigation} & SpatialUnderstanding & 23.36 & 23.90 & 23.82 & 25.77 & 27.68 & 30.15 & 31.77 & 31.72 & 35.86 & 34.43 & 35.14 & 38.52 & 38.54 \\
 & MazeNav & 16.32 & 36.99 & 40.42 & 43.52 & 49.80 & 38.36 & 42.94 & 45.15 & 42.44 & 41.59 & 45.70 & 51.45 & 50.73 \\
 & bAbI(path finding) & 1.48 & 6.81 & 5.20 & 7.14 & 5.75 & 8.44 & 10.20 & 8.99 & 6.66 & 8.42 & 12.01 & 7.51 & 11.36 \\
 & BBH(navigate) & 56.12 & 53.16 & 60.92 & 69.18 & 71.02 & 70.41 & 70.00 & 75.10 & 72.65 & 74.59 & 79.18 & 78.16 & 77.76 \\
\midrule
\multirow{2}{*}{Geometry} & BBH(geometric shapes) & 24.49 & 27.15 & 30.53 & 34.84 & 49.18 & 40.88 & 51.13 & 35.66 & 30.74 & 41.70 & 56.25 & 42.01 & 70.80 \\
 & Geometry3K & 14.56 & 22.53 & 24.41 & 23.41 & 24.83 & 25.25 & 24.16 & 24.12 & 21.90 & 28.31 & 26.89 & 29.36 & 25.46 \\
\midrule
Physical Reasoning & PIQA-spatial & 38.32 & 51.41 & 55.11 & 54.24 & 54.24 & 56.80 & 58.07 & 56.66 & 54.79 & 56.11 & 54.65 & 51.41 & 61.18 \\
\midrule
Symbolic Planning & Blocksworld & 3.26 & 6.31 & 10.44 & 10.64 & 12.98 & 7.54 & 22.05 & 14.41 & 14.36 & 21.44 & 10.90 & 21.84 & 27.60 \\
\midrule
\rowcolor{blue!5} \multicolumn{2}{c|}{Average} & 22.56 & 28.81 & 31.32 & 32.90 & 35.94 & 34.59 & 38.37 & 35.57 & 34.89 & 38.24 & 38.15 & 38.61 & 44.25 \\
\bottomrule\end{tabular}%
\end{table}

\begin{table}[!tb]\centering\scriptsize
\setlength\tabcolsep{6pt}
\caption{\textbf{\textsc{img} spatial scores across various $N$.}}
\label{table:spatial_img}
\begin{tabular}{l|l|cccccc}
\toprule
 & Model Sizes ($N$) & A71M & A128M & A340M & A590M & A874M & A3B \\
\midrule
\multirow{4}{*}{Spatial Relations} & StepGame & 15.46 & 14.20 & 16.97 & 20.64 & 21.67 & 33.11 \\
 & SpaRTUN & 19.76 & 24.11 & 24.95 & 25.17 & 24.68 & 32.17 \\
 & bAbI(positional reasoning) & 28.51 & 48.09 & 47.73 & 47.77 & 47.73 & 58.41 \\
 & FoREST & 34.23 & 35.70 & 40.04 & 40.90 & 47.75 & 52.41 \\
\midrule
\multirow{3}{*}{Localization} & SpartQA & 25.93 & 33.35 & 32.72 & 31.01 & 30.58 & 31.78 \\
 & SpatialMap & 8.78 & 18.61 & 22.19 & 24.35 & 27.63 & 33.25 \\
 & BBH(colored objects) & 12.55 & 12.45 & 14.49 & 21.94 & 25.20 & 54.90 \\
\midrule
\multirow{4}{*}{Navigation} & SpatialUnderstanding & 14.64 & 19.05 & 17.52 & 27.15 & 22.37 & 38.54 \\
 & MazeNav & 11.59 & 22.35 & 24.12 & 26.70 & 32.73 & 50.73 \\
 & bAbI(path finding) & 7.64 & 6.93 & 8.52 & 9.52 & 6.83 & 11.36 \\
 & BBH(navigate) & 46.12 & 53.27 & 53.47 & 60.10 & 66.12 & 77.76 \\
\midrule
\multirow{2}{*}{Geometry} & BBH(geometric shapes) & 22.23 & 29.00 & 15.47 & 12.70 & 30.23 & 70.80 \\
 & Geometry3K & 18.04 & 12.75 & 16.90 & 21.77 & 21.52 & 25.46 \\
\midrule
Physical Reasoning & PIQA-spatial & 11.91 & 46.94 & 51.87 & 51.23 & 52.19 & 61.18 \\
\midrule
Symbolic Planning & Blocksworld & 0.15 & 0.46 & 0.51 & 10.08 & 8.30 & 27.60 \\
\midrule
\rowcolor{blue!5} \multicolumn{2}{c|}{Average} & 15.41 & 24.28 & 25.01 & 28.13 & 30.27 & 44.25 \\
\bottomrule\end{tabular}%
\end{table}

\clearpage

\begin{table}[!tb]\centering\scriptsize
\setlength\tabcolsep{3.6pt}
\caption{\textbf{A3B \textsc{pad} spatial scores across various $D$.}}
\label{table:pt_spatial_pad}
\begin{tabular}{l|l|ccccccccccccc}
\toprule
 & \# Pre-Trained Tokens & 25B & 50B & 75B & 100B & 125B & 150B & 175B & 200B & 225B & 250B & 275B & 300B & 325B \\
\midrule
\multirow{4}{*}{Spatial Relations} & StepGame & 15.72 & 19.23 & 20.70 & 21.46 & 26.37 & 27.28 & 26.95 & 27.72 & 29.63 & 31.77 & 31.73 & 29.41 & 30.02 \\
 & SpaRTUN & 21.41 & 25.37 & 25.14 & 26.45 & 28.83 & 29.83 & 30.60 & 30.49 & 31.64 & 31.95 & 32.33 & 31.20 & 33.32 \\
 & bAbI(positional reasoning) & 46.53 & 47.81 & 47.89 & 50.32 & 48.41 & 51.59 & 52.43 & 49.96 & 52.07 & 50.48 & 52.19 & 57.50 & 52.75 \\
 & FoREST & 33.95 & 46.01 & 46.95 & 44.89 & 49.06 & 51.50 & 48.79 & 48.21 & 51.64 & 50.10 & 49.41 & 49.00 & 48.69 \\
\midrule
\multirow{3}{*}{Localization} & SpartQA & 31.23 & 31.96 & 29.11 & 30.20 & 31.68 & 29.00 & 30.20 & 29.53 & 30.12 & 29.35 & 28.84 & 30.03 & 28.97 \\
 & SpatialMap & 16.22 & 22.77 & 28.18 & 25.95 & 30.07 & 31.04 & 30.62 & 30.44 & 27.76 & 31.19 & 30.74 & 34.05 & 31.92 \\
 & BBH(colored objects) & 15.20 & 18.78 & 20.92 & 25.92 & 23.88 & 29.90 & 33.98 & 36.53 & 36.73 & 37.65 & 34.80 & 42.24 & 39.59 \\
\midrule
\multirow{4}{*}{Navigation} & SpatialUnderstanding & 20.88 & 23.08 & 27.38 & 26.79 & 23.77 & 26.86 & 28.47 & 31.98 & 33.45 & 30.87 & 32.02 & 33.69 & 32.40 \\
 & MazeNav & 15.54 & 31.13 & 44.87 & 38.94 & 38.61 & 35.24 & 23.05 & 38.46 & 34.91 & 35.66 & 32.36 & 46.40 & 46.35 \\
 & bAbI(path finding) & 6.51 & 4.12 & 2.96 & 4.12 & 7.36 & 7.89 & 10.60 & 8.54 & 10.13 & 7.54 & 15.88 & 12.96 & 12.41 \\
 & BBH(navigate) & 51.22 & 51.53 & 59.90 & 64.49 & 63.47 & 68.88 & 66.63 & 67.65 & 73.37 & 65.51 & 67.24 & 76.53 & 76.73 \\
\midrule
\multirow{2}{*}{Geometry} & BBH(geometric shapes) & 8.30 & 27.87 & 32.38 & 39.75 & 57.58 & 48.05 & 42.11 & 52.15 & 44.47 & 42.93 & 45.49 & 48.67 & 60.35 \\
 & Geometry3K & 12.75 & 15.10 & 18.04 & 22.27 & 21.94 & 24.33 & 20.64 & 22.15 & 21.77 & 26.17 & 28.23 & 24.37 & 27.68 \\
\midrule
Physical Reasoning & PIQA-spatial & 38.73 & 51.64 & 53.06 & 52.46 & 53.70 & 57.39 & 56.48 & 57.30 & 56.25 & 58.62 & 58.58 & 55.29 & 57.53 \\
\midrule
Symbolic Planning & Blocksworld & 1.78 & 3.51 & 8.20 & 6.92 & 4.79 & 10.34 & 12.27 & 11.30 & 9.57 & 10.18 & 10.59 & 5.86 & 21.03 \\
\midrule
\rowcolor{blue!5} \multicolumn{2}{c|}{Average} & 20.81 & 27.20 & 30.25 & 31.19 & 33.04 & 34.78 & 33.93 & 35.61 & 34.95 & 35.34 & 35.96 & 36.21 & 39.87 \\
\bottomrule\end{tabular}%
\end{table}

\begin{table}[!tb]\centering\scriptsize
\setlength\tabcolsep{6pt}
\caption{\textbf{\textsc{pad} spatial scores across various $N$.}}
\label{table:spatial_pad}
\begin{tabular}{l|l|cccccc}
\toprule
 & Model Sizes ($N$) & A71M & A128M & A340M & A590M & A874M & A3B \\
\midrule
\multirow{4}{*}{Spatial Relations} & StepGame & 13.50 & 13.51 & 17.14 & 20.94 & 22.34 & 30.02 \\
 & SpaRTUN & 16.03 & 19.69 & 25.82 & 24.59 & 25.98 & 33.32 \\
 & bAbI(positional reasoning) & 47.33 & 47.05 & 48.37 & 48.41 & 48.92 & 52.75 \\
 & FoREST & 27.69 & 36.37 & 45.08 & 42.93 & 47.61 & 48.69 \\
\midrule
\multirow{3}{*}{Localization} & SpartQA & 22.81 & 26.62 & 31.80 & 31.03 & 32.33 & 28.97 \\
 & SpatialMap & 8.32 & 18.33 & 18.88 & 24.85 & 26.63 & 31.92 \\
 & BBH(colored objects) & 7.04 & 14.18 & 19.80 & 19.08 & 23.27 & 39.59 \\
\midrule
\multirow{4}{*}{Navigation} & SpatialUnderstanding & 13.07 & 17.67 & 20.42 & 21.84 & 25.41 & 32.40 \\
 & MazeNav & 11.71 & 17.77 & 21.25 & 21.30 & 24.47 & 46.35 \\
 & bAbI(path finding) & 4.40 & 5.15 & 6.98 & 6.73 & 11.11 & 12.41 \\
 & BBH(navigate) & 33.06 & 45.51 & 58.37 & 58.57 & 59.29 & 76.73 \\
\midrule
\multirow{2}{*}{Geometry} & BBH(geometric shapes) & 18.44 & 11.68 & 15.27 & 14.96 & 19.26 & 60.35 \\
 & Geometry3K & 6.25 & 9.48 & 12.88 & 19.92 & 21.77 & 27.68 \\
\midrule
Physical Reasoning & PIQA-spatial & 23.68 & 38.91 & 51.00 & 50.87 & 51.78 & 57.53 \\
\midrule
Symbolic Planning & Blocksworld & 0.15 & 4.07 & 0.76 & 7.18 & 8.30 & 21.03 \\
\midrule
\rowcolor{blue!5} \multicolumn{2}{c|}{Average} & 15.10 & 20.66 & 25.03 & 26.97 & 29.05 & 39.87 \\
\bottomrule\end{tabular}%
\end{table}

\end{document}